\documentclass[11pt]{article}
\usepackage[margin=0.8in]{geometry}
\usepackage[ruled,vlined]{algorithm2e}
\usepackage{graphicx,subfigure}
\usepackage{amssymb}
\usepackage{amsthm}
\usepackage{comment}
\usepackage{natbib}

\newcommand{\RNum}[1]{\mathrm{\uppercase\expandafter{\romannumeral #1\relax}}}
\usepackage{enumitem} 
\usepackage{amsmath, amssymb}
\usepackage{xcolor}

\usepackage[utf8]{inputenc} 
\usepackage[T1]{fontenc}  
\usepackage{hyperref}    
\usepackage{url}      
\usepackage{booktabs}    
\usepackage{amsfonts}    
\usepackage{nicefrac}    
\usepackage{multirow}
\usepackage{float}
\usepackage{microtype}   
\usepackage{xcolor}     
\usepackage[symbol]{footmisc}
\renewcommand{\thefootnote}{\fnsymbol{footnote}}

\makeatletter
\newcommand*{\rom}[1]{\expandafter\@slowromancap\romannumeral #1@}
\makeatother 

\allowdisplaybreaks

\begin{document}
\renewcommand{\thefootnote}{\arabic{footnote}}
    \title{\textbf{Are machine learning interpretations reliable?}\\  A stability study on global interpretations}
	\author{Luqin Gan$^1$, Tarek M. Zikry$^{2,3}$, Genevera I. Allen$^{2,3,4}$}
	\date{}
	\maketitle
\begin{abstract}
As machine learning systems are increasingly used in high-stakes domains, there is a growing emphasis placed on making them interpretable to improve trust in these systems. In response, a range of interpretable machine learning (IML) methods have been developed to generate human-understandable insights into otherwise black box models. With these methods, a fundamental question arises: \textit{Are these interpretations reliable?} Unlike with prediction accuracy or other evaluation metrics for supervised models, the proximity to the true interpretation is difficult to define. Instead, we ask a closely related question that we argue is a prerequisite for reliability: \textit{Are these interpretations stable?} We define stability as findings that are consistent or reliable under small random perturbations to the data or algorithms.  In this study, we conduct the first systematic, large-scale empirical stability study on popular machine learning global interpretations for both supervised and unsupervised tasks on tabular data. Our findings reveal that popular interpretation methods are frequently unstable, notably less stable than the predictions themselves, and that there is no association between the accuracy of machine learning predictions and the stability of their associated interpretations. Moreover, we show that no single method consistently provides the most stable interpretations across a range of benchmark datasets. Overall, these results suggest that interpretability alone does not warrant trust, and underscores the need for rigorous evaluation of interpretation stability in future work. To support these principles, we have developed and released an open source IML dashboard and Python package to enable researchers to assess the stability and reliability of their own data-driven interpretations and discoveries.

\end{abstract}

	\footnotetext[1]{Department of Statistics, Rice University,
Houston, TX}

	\footnotetext[2]{Departments of Statistics, Columbia University, New York, NY}
    \footnotetext[3]{Center for Theoretical Neuroscience, Zuckerman Mind Brain Behavior Institute, Columbia University, New York NY}
    \footnotetext[4]{\url{genevera.allen@columbia.edu}}
\renewcommand{\thefootnote}{\fnsymbol{footnote}}

\section{Introduction}
In recent years the field of Interpretable Machine Learning (IML) has arisen, a class of ML methods focused "the use of machine learning techniques to generate human understandable insights into data, the learned model, or the model output" \citep{allen2023interpretable}. These methods are employed to generate data-driven discoveries in both unsupervised and supervised settings, with a focus on opening up the black box of machine learning to users and practitioners across domains, particularly critical in domains such as health, security, and even finance, where ML methods have been applied to approve or deny mortgages \citep{van2022increasing}. Existing review papers have defined concepts of IML methods, provided systematic taxonomy and principles \citep{yu2020veridical}, and discussed its importance in various aspects including model validation 
\cite{du2019techniques,murdoch2019definitions,bastani2017interpretability}, model debugging \cite{molnar2019,du2019techniques}, model transparency \& accountability \cite{du2019techniques, lipton2018mythos,guidotti2018survey,tomsett2018interpretable,carvalho2019machine}, trust \cite{ignatiev2020towards,ribeiro2016should, siau2018building, toreini2020relationship, drozdal2020trust,rossi2018building,ribeiro2016should}, ethics \cite{molnar2019, doshi2017towards,lipton2018mythos,doshi2018considerations,watson2021explanation,carvalho2019machine}, and scientific discoveries \cite{roscher2020explainable,molnar2019,du2019techniques,lipton2018mythos,doshi2018considerations,watson2021explanation, zikry2025limeade}. 
Though these terms have been defined before, we redefine them here for clarity:
\begin{itemize}
    \item \textbf{Reproducibility:} Do you observe the same results when the pipeline is rerun on the same data?
    \item \textbf{Replicability:} Is your code and pipeline easily usable to run on unseen data?
    \item \textbf{Reliability:}  Do you observe similar results to the ground truth?
    \item \textbf{Generalizability/Predictability:} Can you predict your results or observe similar results on unseen data?
    \item \textbf{Stability:} How sensitive are your results to random data or algorithm perturbations? 
\end{itemize}

Here, we believe a broad prerequisite criterion to trust interpretations is \textit{reliability}. For example, from the perspective of knowledge discovery, a new finding cannnot be accepted if the discovery is unreliable after many repeated experiments. Therefore, we consider the reliability of the interpretations to be significant in building trustworthy AI systems. As machine learning methods become ever present in a wide range of critical domains, including biomedicine and drug discovery \citep{goecks2020machine}, physics \citep{carleo2019machine}, national security \citep{haney2020applied}, finance \citep{dixon2020machine}, climate science \citep{monteleoni2013climate}, transportation \citep{zantalis2019review}, and more \cite{vilone2020explainable,adadi2018peeking}, significant attention has been drawn towards the quality of interpretations generated by ML models. In supervised learning tasks such as regression and classification, interpretations include feature importance ranking for a predictive model, whereas in an unsupervised task such as clustering and dimension reduction, interpretations are in the groupings generated, either in the high-dimensional space or a low-dimensional embedding. These interpretations are widely used across domains, either as standalone findings such as ranking features important in aircraft engine longevity \citep{alomari2024shap}, or as part of step in a downstream analysis, such as identifying groups of drug-resistant cancer cells \citep{zikry2024cell}. 


Though vast bodies of work have been devoted to ML methods, and recently, IML methods, the reliability of interpretations themselves have not been deeply explored. As compared to traditional statistical tasks such as prediction, in which we can easily measure its accuracy with the true response, the quality of interpretations from a machine learning model are more difficult to evaluate, as ground truth interpretations are rarely observed, making measures of reliability challenging. In lieu of being able to directly compare to ground truth interpretations, many have turned to studying \textit{stability} as a practical proxy; if an interpretation changes dramatically under small perturbations to the data or model, it is unlikely to be reliable. Since the inception of the idea to study feature selection in regularized models \citep{meinshausen2010stability}, further work has arisen for feature interactions \cite{basu2018iterative, little2025iloco}, graphical models \citep{liu2010stability, muller2016generalized}, and dimension reduction \citep{xia2024statistical, campbell2015laplacian}. In clustering, it is commonly referred to as consensus clustering \citep{monti2003consensus}, where it has gained popularity for identifying optimal hyperparameters in unsupervised class discovery \citep{wu2014k, xanthopoulos2014review, lock2013bayesian}. Though widely advocated for validating reproducible discovery \citep{allen2023interpretable, yu2020veridical}, stability analyses can quickly become computationally burdensome, and the choice of perturbation (noise addition, different train/test splits, random reinitializations) and evaluation metrics are not always immediately clear, and can require specific methodological knowledge \cite{xu2018splitting, allen2023interpretable}. 

Despite the large volume of IML-related research on supervised learning methods \cite{du2019techniques,rudin2019stop,doshi2017towards,gilpin2018explaining,tomsett2018interpretable}, there is no systematic framework to evaluate the reliability of the interpretations on unsupervised scientific discovery. \cite{liu2020impact} explored the impact of prediction accuracy on the quality interpretability in terms of accuracy and stability of feature importance ranking. However, this work focuses only on feature importance and intrinsic global IML methods. Many past efforts focus on the reliability of \textit{model-specific} methods, such as \cite{zhang2019should} who investigated the uncertainties in the surrogate interpretability model LIME \cite{ribeiro2016should} due to sampling procedure, and sampling proximity and explained model credibility across different observations. Similarly, \cite{adebayo2018sanity} provides a framework to evaluate the adequacy of explanations obtained from saliency map-related methods, which measures the sensitivity of explanations to label permutation or random parameters. As a useful framework for model validation, however, it is limited to saliency maps and provides no insights into the reliability of the explanations. Additional literature explored the sensitivity of explanations by parameter randomization or feature perturbation, primarily in deep learning methods \cite{adebayo2018local,adebayo2018sanity,ancona2017towards}. Past taxonomies classified the evaluation of interpretability into application-grounded, human-grounded, and functionally-grounded evaluation \cite{doshi2017towards}. 

\paragraph{Contribution} In this study, we carry out a wide-ranging systematic empirical assessment of the stability of commonly used global interpretable machine learning (IML) methods on tabular data.  We study the stability of interpretations via data perturbations including train/test splits and noise addition. Specifically, we will focus on issues of the stability of interpretations across four main tasks: 1) classification, 2), regression, 3) clustering, and 4) dimensionality reduction. To facilitate transparency and reproducibility, we introduce an interactive, open-source software framework that enables practitioners to explore and visualize the reliability of IML outputs across diverse datasets and modeling scenarios. By combining these empirical investigations with user-friendly tooling, we aim to deliver critical insights and practical guidance on how much we can trust machine-learning interpretations in high-stakes domains such as human health and society.

\section{Study Description}
\begin{figure*}[tbhp]
\centering
\includegraphics[width=0.8\linewidth]{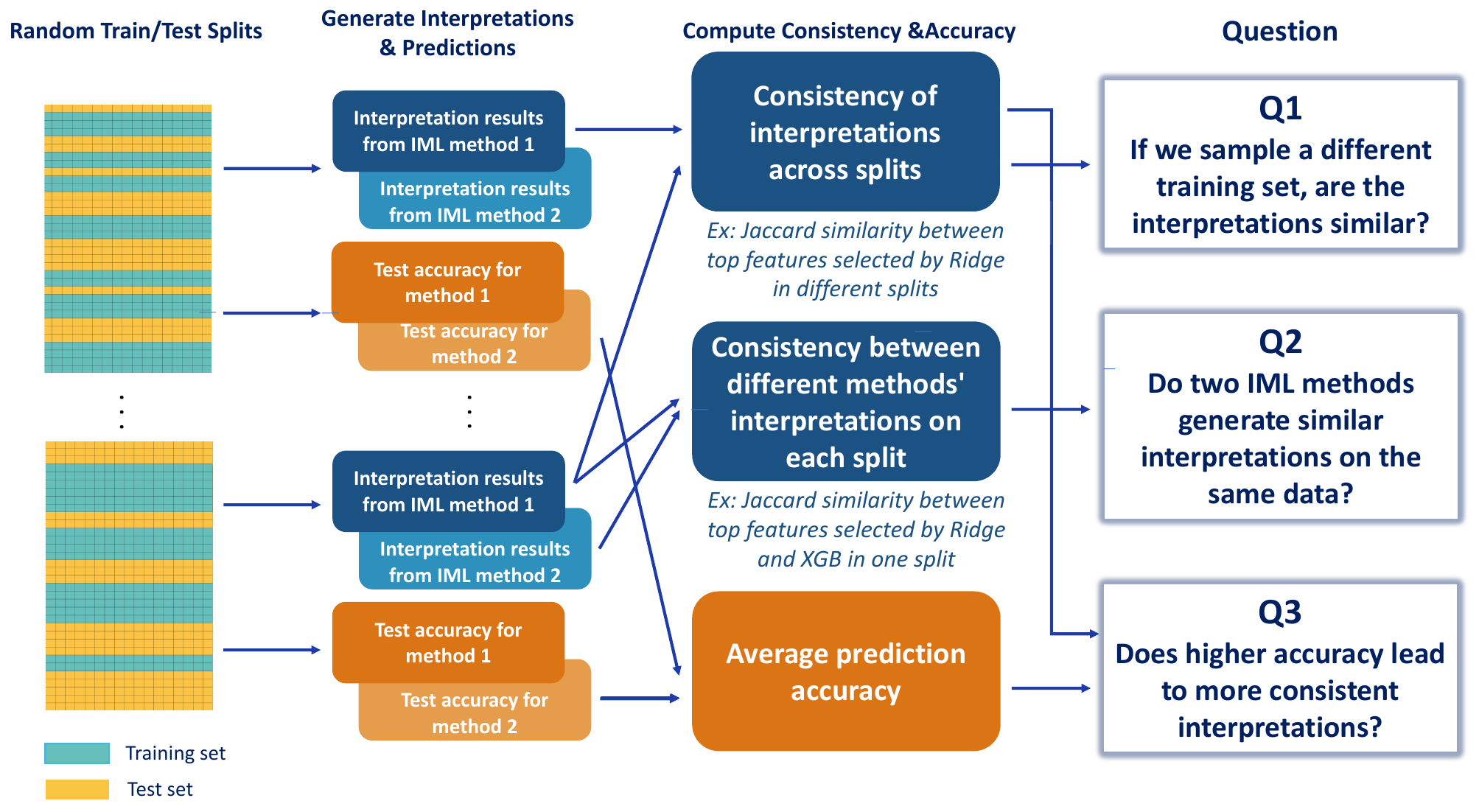}
\caption{Overview of study design with data splitting. After conducting multiple random data splitting, interpretations, and test predictions are generated using different IML methods on each training set. Then the within-method stability and between-method stability, and average prediction accuracy on the test set are computed. }
\label{fig:f1}
\end{figure*}

As detailed in Figure~\ref{fig:f1}, we seek to create a wide-ranging empirical framework to answer three key questions on the reliability of interpretations in machine learning methods across four tasks 1) classification, 2) regression, 3) clustering, and 4) dimension reduction. Through publicly available benchmarking datasets, we apply commonly used methods in each of these four tasks. We evaluate a diverse set of IML methods spannining multiple paradigms including comparisons between instrinsic and post-hoc approaches, model-specific and model-agnostic approaches, and global and local methods of interpretability. We refer readers to \citep{allen2023interpretable, du2019techniques,doshi2017towards,guidotti2018survey} for further clarifications on these categorizations.

\paragraph{Scope} 
For data types, we focus on tabular data only, hence the literature around IML for image, textual, or sequential data-related methods are not included. We focus on evaluating the reliability of interpretations obtained from three major types of machine learning tasks, including feature importance in supervised learning (regression and classification), clustering, and dimension reduction in unsupervised learning. We evaluate a wide range of popular model-specific and model-agnostic, intrinsic and post-hoc methods. However, we only consider global interpretation methods or local methods which can be aggregated to global representation such as Layer-wise Relevance Propagation (LRP). For the type of explanations, we consider quantiative explanations only, excluding rule-based or visual interpretations.

We focus on the most popular machine learning and deep learning methods in these areas, and compute metrics to measure the reliability of the interpretations through empirical analysis to a wide range of publicly-available tabular data sets. Since we also aim to test the interpretations derived from generic global methods against each other, we do not apply posthoc methods which are specifically applied to certain machine learning models for convolutional neural network, or methods that provide local feature importance, such as LIME \citep{ribeiro2016should} or anchor \citep{ribeiro2018anchors}. Specifically, we include general model-specific machine learning models, deep learning-related methods, and model-agnostic methods. Further details on methods and implementation can be found in Appendix \ref{app:studydesign}. Through this analysis, we seek to answer three fundamental questions in IML:
\subsection{Questions}
\paragraph{Q1: If we sample a different training set, are the interpretations similar?} 
Within-method stability reflects how consistently an interpretation method performs when the input data is slightly altered. If small perturbations, such as resampling via different train/test splits, adding noise, or permuting features, lead to substantially different interpretations, the IML method cannot be considered stable or trustworthy for scientific discovery. Interpretations that are highly sensitive to these minor changes, or conditional on a particular train/test split, risk producing conclusions that are not robust or reproducible, and may mislead downstream analyses or decision-making.

\paragraph{Q2: Do two IML methods generate similar interpretations on the same data?}
This question seeks to compare between-method stability. On the same data, different practitioners often turn to different methods to answer the same questions in IML. However, if these methods lead to different interpretations, how do we know which interpretation is trustworthy? If these methods yield different interpretations, it raises the question of which interpretation, if any, can be trusted, emphasizing the need for a systematic evaluation of their consistency and validity to ensure reliable conclusions. 

\paragraph{Q3: Does higher accuracy lead to more consistent interpretations?}
Past work has suggested that predictive accuracy is highly relevant to interpretation stability \cite{murdoch2019interpretable,liu2020impact}. To address this, we also explore the relationship between interpretation stability and prediction accuracy in greater depth to determine if accuracy can be used as a determining factor for selecting an IML method.

\subsection{Machine Learning Tasks}
Most of the existing literature on interpretable machine learning techniques is focused on the interpretations of feature importance from supervised learning methods, or on model-specific metrics of reliability. The interpretability, particularly stability, of unsupervised learning methods, remains understudied in the literature \citep{vilone2020explainable}. Therefore, we focus on evaluating the reliability of interpretations obtained from three major types of machine learning tasks, including feature importance in supervised learning, clustering, and dimension reduction in unsupervised learning. We focus on \textit{global} methods, or when using local methods, we sum up over all points in the training set to obtain global interpretations.
\subsubsection{Classification \& Regression}
\begin{table}[h!]
\centering
{\begin{tabular}{lrrr}
\toprule

Method & Local/Global & Model Specific/Model Agnostic & Intrinsic/Post-hoc\\
\midrule
LASSO & Global & Model Specific &Intrinsic \\
Ridge & Global & Model Specific &Intrinsic\\
Decision Tree & Global & Model Specific&Intrinsic\\
Random Forest& Global & Model Specific &Intrinsic\\
XGBoost& Global & Model Specific &Intrinsic\\

Linear SVM & Global & Model Specific&Intrinsic \\
deepLIFT & Local & Model Specific &Post-hoc\\
Guided Backpropagation & Local & Model Specific &Post-hoc\\
Epsilon-LRP & Local & Model Specific &Post-hoc\\ 
Saliency maps  & Local & Model Specific &Post-hoc\\
Integrated Gradients & Local & Model Specific &Post-hoc\\
Occlusion & Local & Model Agnostic &Post-hoc\\

Permutation  & Global & Model Agnostic &Post-hoc\\
Shapley value   & Local & Model Agnostic &Post-hoc\\  
\bottomrule
\end{tabular}}
\caption{Feature Importance in Classification/Regression Methods Summary.}
\label{table:featimp}
\end{table}

Most IML review papers focus on the discussion of techniques offering interpretations of feature importance \citep{du2019techniques,murdoch2019definitions,bastani2017interpretability}. Such methods aim to provide information on the level of each feature's contribution to predictions by assigning an importance score or ranking to individual features by their level of relevance. Feature importance techniques present a \textit{global} relationship between each feature and the outcome. Intrinsic and \textit{model-specific} feature importance methods include commonly used statistical methods such as generalized linear model, naive Bayes classifier, SVM, and tree-based gradient boosting and random forest. For example, coefficients in generalized linear models and information gain from tree-based methods\citep{scikit-learn} are model-specific interpretations. Secondly, numerous deep learning-specific IML techniques have been proposed to untangle the interpretability challenges of the black box of neural networks. 
Other lines of work measures features importance using the changes of prediction outcome after intentional feature occlusion \citep{li2016deep}, perturbation \citep{fong2017interpretable} or adversarial change \citep{papernot2018deep,baehrens2010explain, murdoch2019definitions}. For \textit{model-agnostic} methods, some use counterfactual explanations \citep{verma2020counterfactual} of a prediction to evaluate feature importance, including partial dependent plot (\textit{global}), accumulated local effects plot (\textit{global}), individual conditional expectation plot (\textit{local}). Another line of work measures features importance using the changes of prediction outcome after intentional feature occlusion (\textit{local}) \citep{li2016deep}, perturbation (\textit{global}) \citep{fong2017interpretable}, Shapley value (\textit{local}) \citep{lundberg2017unified}, LIME (\textit{local}) \citep{ribeiro2016should} or adversarial change \citep{papernot2018deep,baehrens2010explain, murdoch2019definitions}. For this study, we choose only global methods, or local methods that can be aggregated into global metrics, summarized in Table~\ref{table:featimp}. For local methods, we sum the metrics over all points in the training set to obtain global interpretations.

\subsubsection{Dimensionality Reduction}
\begin{table}[h!]
\centering
{\begin{tabular}{lrr}
\toprule

Dimension Reduction Method & Global/Local & Linear/Non-Linear\\
\midrule
PCA & Global &Linear \\
Random projection& Global &Linear \\
Metric MDS (euclidean)& Global &Non-linear \\
Non-metric MDS & Local &Non-linear \\
Isomap & Global &Non-linear \\
Deep Autoencoder & --- &Non-linear \\
tSNE& Local &Non-linear \\
UMAP& Local &Non-linear \\
Spectral Embedding& Global &Non-linear \\

\bottomrule
\end{tabular}}
\caption{Dimension Reduction Methods Summary.}
\label{table:dr}
\end{table}
Dimension reduction (DR) methods aim to project the original data into lower dimensions that reflect latent trends. In the high dimensional setting, the original data might be redundant with noisy features or highly correlated features \citep{carreira1997review}, which would significantly hamper the prediction performance of machine learning methods. Dimension reduction methods overcome the curse of dimensionality by transforming such complex data with a large number of covariates to lower dimensions while preserving the information contained. The main usage of dimension reduction techniques include (1) to draw inferences on the data structure by visualization \cite{jung2009pca,rudin2022interpretable}, which implies relative distances among the observations in the reduced space, (2) to be applied as pre-processing step for further downstream analysis clustering on the reduced embedding \cite{carreira1997review}. The interpretation task of DR methods is to (1) retain the neighbor structure of the original data in lower dimension space; (2) recover correct cluster labels by applying clustering on reduced dimension. 

Our choices of dimension reduction methods and their categorizations can be seen in Table~\ref{table:dr}. We select the most widely used linear global technique: principal component analysis (PCA), and include random projections as a baseline comparison. For non-linear manifold learning methods, we select methods that preserve \textit{global} properties such as metric/non-metric multidimensional scaling (MDS) \cite{davison2000multidimensional}and Isomap \cite{balasubramanian2002isomap}, and methods that preserve \textit{local} properties including T-distributed Stochastic Neighbor Embedding (tSNE) \citep{van2008visualizing}, Uniform Manifold Approximation and Projection for Dimension Reduction (UMAP) \citep{mcinnes2018umap}, spectral embedding with different affinity \citep{shi2000normalized}. In addition, we apply a deep auto-encoder framework \citep{kramer1991nonlinear} with a three-layered encoder and a three-layered decoder, and the low-dimensional code produced by the encoder is used as the reduced dimensions.

\subsubsection{Clustering}
\begin{table}[h!]
\centering
\small
\begin{tabular}{llrrr}
\toprule
\bf Method & \bf Dataset & \bf Complete Linkage & \bf Average Linkage & \bf Single Linkage \\
\midrule
\multirow{14}{*}{\textbf{Hierarchical}} 
 & Bean & canberra & canberra & euclidean \\
 & Call & euclidean & cosine & manhattan \\
 & Statlog & cosine & cosine & chebyshev \\
 & Spambase & canberra & cosine & chebyshev \\
 & Iris & canberra & cosine & chebyshev \\
 & WDBC & canberra & chebyshev & euclidean \\
 & Tetragonula & manhattan & manhattan & manhattan \\
 & Author & cosine & canberra & manhattan \\
 & Ceramic & canberra & canberra & euclidean \\
 & TCGA & euclidean & canberra & canberra \\
 & Psychiatrist & euclidean & euclidean & euclidean \\
 & Veronica & canberra & euclidean & euclidean \\
 & Asian Religions & canberra & canberra & euclidean \\
 & PANCAN & canberra & canberra & chebyshev \\
\midrule
\multicolumn{2}{l}{\textbf{k-Means}} \\
 & k-Means++  \\
 & MiniBatch k-Means \\
\midrule
\multicolumn{2}{l}{\textbf{Spectral Clustering}} \\
 & Nearest Neighbors Affinity  \\
 & RBF Kernel Affinity  \\
\midrule
\textbf{BIRCH}  \\
\bottomrule
\end{tabular}
\caption{Summary of best-performing distance metrics for hierarchical clustering, and hyperparameter variants used for other clustering methods.}
\label{table:dis}
\end{table}
Clustering methods identify partitions among the data, subsets of observations sharing some key characteristics.  Clustering models are \textit{intrinsic} and \textit{model-specific}, which provide the discovery of groups by generating a set of homogeneous subgroups with respect to the hidden data structure. Traditional clustering techniques, such as K-Means and hierarchical clustering, aim partition observations into disjoint groups \citep{Hennig2015HandbookOC}, with broad applications in bioinformatics, computational biology, and character recognition. Recently, neural network-based clustering techniques are increasingly developed \citep{min2018survey}. The deep learning methods are particularly powerful for large data sets, utilizing architectures such as autoencoders or restricted Boltzmann machines \citep{min2018survey,huang2019unsupervised}. On the other hand, overlapping clusters such as fuzzy clustering can assign observations into multiple groups, which are more helpful in the application to social network detection \citep{rokach2010survey}. There are well-established theoretic foundations for the stability of K-means clustering \citep{pollard1981strong}, spectral clustering \citep{von2008consistency,trillos2018variational}. However, none have studied the empirical stability of partitions obtained from clustering methods. Table!\ref{table:dis} summarizes our clustering method and hyperaparameters of choice by dataset.

\subsection{Reliability Test and Metrics}

\paragraph{Reliability test}\label{sec:test}
The first step of our framework is to design sensitivity tests to obtain interpretations from machine learning models under different circumstances. We propose two types of perturbation techniques to conduct sensitivity tests: random splits and adding random noise. In supervised learning, researchers conduct a train/test split before fitting a predictive model so as to avoid overfitting. The interpretations are reliable if the resulting feature importance scores are consistent with different random train/test splits. We also apply data splitting for unsupervised tasks and fit IML methods to the training sets. We add random noise for unsupervised learning tasks including clustering and dimension reduction methods. The reliability of an IML method can be measured by the consistency after performing multiple data perturbations. Specifically, we first perform random perturbation on the data set of interest, by either conducting train/test splits, or adding random noise. To carry out this empirical study, we construct a custom web app to explore the reliability of ML interpretations, available at \url{https://iml-reliability.herokuapp.com/home}.

\paragraph{Stability metrics}\label{sec:metric}


\begin{table}[h!]
\centering
\small 
\renewcommand{\arraystretch}{1.5} 
\begin{tabular}{lllcp{8cm}}
\toprule
\bf Metric & \bf Task(s) & \bf Equation & \bf Citation \\
\midrule
Jaccard Similarity & C1, R & $J(A,B)@k =\frac{|A_k\cap B_k|}{|A_k\cup B_k|}$ & \cite{real1996probabilistic} \\

AO (Adjusted Overlap) & C1, R & $AO@k  = \frac{1}{k} \sum_{d=1}^{k} \frac{|A_d\cap B_d|}{d}$ & \cite{webber2010similarity} \\

Top K Kendall's Tau & C1, R & $K(A,B)^{(p)}@k =\sum_{i,j \in \mathcal{P}(A_k,B_k)} \bar{K}_{i,j} ^{(p)}(A_k,B_k)$ & \cite{kendall1938new} \\

ARI (Adjusted Rand Index) & C2, D & $ARI  = \frac{\sum_{ij} {n_{ij}\choose 2} - [\sum_{i} {a_{i}\choose 2} \sum_{j} {a_{j}\choose 2}]/ {n\choose 2}}{\frac{1}{2}[\sum_{i} {a_{i}\choose 2} + \sum_{j} {a_{j}\choose 2}] - [\sum_{i} {a_{i}\choose 2} \sum_{j} {a_{j}\choose 2}]/ {n\choose 2}}$ & \citep{rand1971objective} \\

Fowlkes-Mallows Index & C2, D & $FM = \sqrt{PPV \cdot TPR}$ &\cite{fowlkes1983method} \\

Mutual Information & C2, D & $MI(A,B)  = \sum_{i\in A}\sum_{j\in B} p(i,j)\log\frac{p(i,j)}{p(i)p(j)}$ & \cite{kraskov2004estimating} \\

V-measure & C2, D & $V = \frac{(1+\beta)\cdot\text{homogeneity}\cdot\text{completeness}}{\beta \cdot \text{completeness}+\text{homogeneity}}$ & \cite{rosenberg2007v} \\

NN-Jaccard-AUC & D & $\text{NN-J-AUC}(A,B)@k = \frac{1}{N} \sum_{i=1}^{N} \text{AUC}(J(A_i, B_i)@k)$ &  \\

\bottomrule
\end{tabular}
\caption{Stability Metrics Summary. C1 indicates metrics used for classification, C2 indicates clustering, R indicates regression, and D indicates dimension reduction.}
\label{table:fi}
\end{table}
To assess interpretation stability, we implement several robust metrics for each IML task. For feature importance rankings, we measure the stability of top K features by (1) Jaccard similarity \cite{real1996probabilistic}, (2) average overlap (AO) \cite{webber2010similarity} and (3) Kendall Tau distance \cite{kendall1938new}. The stability of clustering labels can be measured by widely used cluster validity indices including adjusted rand index (ARI) \citep{rand1971objective}, mutual information (MI) \cite{kraskov2004estimating}, V measure \cite{rosenberg2007v} and Fowlkes mallows index \cite{fowlkes1983method}. For dimension reduction interpretations, we measure their : (1) clustering stability performed on the reduced dimension, measured by the same metrics as clustering, and (2) visualization reliability measured by local neighbor stability. Here, we propose to examine whether every sample has a consistent set of nearest neighbors from different random perturbations, under the same setting. Specifically, the similarity of two sets of nearest neighbors can be calculated by the Jaccard score \cite{real1996probabilistic}. With the number of nearest neighbors $K$ ranging from 1 to $N$, where $N$ is the total number of samples, we can draw a receiver operating characteristic curve of Jaccard scores against $K$, and obtain the area under the curve ($AUC$) of the Jaccard scores curve. We use the resulting AUC score as a local neighbor stability metric, denoted as \textit{NN-Jaccard-AUC} score. Table \ref{table:fi} shows all stability metrics, and further details are available in the Appendix.

\subsection{Datasets} 
\begin{table}[h!]
\centering
\resizebox{\textwidth}{!}{
\begin{tabular}{lrrrrl}
\toprule
Data & \bf $N$ & \bf $p$ & \bf \# classes & \bf Type & \bf Task(s) \\
\midrule
Online News Popularity \cite{fernandes2015proactive} & 39644 & 59 & - & predict \# of shares in social networks & R \\
BlogFeedback \cite{buza2014feedback} & 52397 & 280 & - & predict how many comments the post will receive & R \\
Satellite image \cite{romano2021pmlb} & 6435 & 36 & - & - & R \\
STAR \cite{DVN/SIWH9F_2008} & 2161 & 39 & - & Tennessee Student Teacher Achievement Ratio (STAR) project & R \\
Communities and crime \cite{blake1998uci} & 1993 & 99 & - & predict \# of violent crimes & R \\
Bike \cite{fanaee2014event} & 731 & 13 & - & hourly and daily count of rental bikes & R \\
CPU \cite{romano2021pmlb} & 209 & 7 & - & - & R \\
Wine \cite{cortez2009modeling} & 178 & 13 & - & red wine quality & R \\
Music \cite{romano2021pmlb} & 1059 & 117 & - & geographical origin of music & R \\
Residential \cite{rafiei2016novel} & 372 & 103 & - & predict house price & R \\
Tecator \cite{romano2021pmlb} & 240 & 124 & - & - & R \\
Word \cite{blake1998uci} & 523 & 526 & - & word occurrence data to predict the length of a newsgroup record & R \\
Riboflavin \cite{buhlmann2014high} & 71 & 4088 & - & genomics data set about riboflavin production rate & R \\
Bean \cite{blake1998uci} & 13611 & 16 & 7 & images of grains of different registered dry beans & C1-2 \\
Call \cite{blake1998uci} & 7195 & 22 & 10 & acoustic features from frog calls & C1-2 \\
Iris \cite{blake1998uci} & 150 & 4 & 3 & types of iris plant & C2 \\
Ceramic \cite{blake1998uci} & 88 & 17 & 2 & chemical composition of ceramic samples & C2 \\
Statlog \cite{blake1998uci} & 2310 & 19 & 7 & image segmentation & C1-2, D \\
Spambase \cite{blake1998uci} & 4601 & 57 & 2 & spam and non-spam email & C1-2, D \\
WDBC \cite{blake1998uci} & 569 & 30 & 2 & breast cancer diagnostic & C2, D \\
Tetragonula \cite{franck2004nest} & 236 & 13 & 9 & tetragonula bee species & C2, D \\
Author \cite{blake1998uci} & 841 & 69 & 4 & word counts from famous authors & C1-2, D \\
TCGA \cite{cancer2012comprehensive} & 445 & 353 & 5 & breast cancer data & C1-2, D \\
Veronica \cite{martinez2004species} & 206 & 583 & 8 & AFLP data of Veronica plants & C2, D \\
Asian Religions \cite{sah2019asian} & 590 & 8266 & 8 & characteristics of Asian religions & C1-2, D \\
PANCAN \cite{weinstein2013cancer} & 761 & 13244 & 5 & high dimensional RNA-seq & C1-2, D \\
\bottomrule
\end{tabular}}
\caption{Datasets for classification, regression, and dimensionality reduction tasks. C1 indicates data used for classification, C2 indicates clustering.}
\label{table:merged}
\end{table}
The study is conducted on 25 popular benchmarking datasets across domains for the three tasks of regression, classification, and dimensionality reduction, characterized in Table \ref{table:merged}. In order to assess the stability of interpretations over a wide range of accuracy scopes, we also ensure to have a variety of ranges of prediction/clustering accuracy for each task. 

Specifically, for the classification task, we include three high-dimension data sets ($P>N$): Asian Religions\citep{sah2019asian}, PANCAN tumor cell \citep{weinstein2013cancer}, and DNase sequencing \citep{encode2012integrated}; six mid-size data sets: Statlog image segment data\citep{blake1998uci}, spam base data set \citep{blake1998uci}, author word counts data set \citep{blake1998uci}, Amphibians species prediction data \cite{habib2020presence}, Madelon artificial data from NIPS 2003 Feature Selection Challenge \citep{guyon2004result}, and breast cancer TCGA data\cite{cancer2012comprehensive}; and four data sets with large $N$ ($N>5000$): bean image attributes data \citep{blake1998uci}, anuran call data \citep{blake1998uci}, theorem prediction data\citep{bridge2014machine} and Digit MNIST data \citep{deng2012mnist}. In the case of regression, we have two high dimensional riboflavin genomics data set \citep{buhlmann2014high} and word occurrence data to predict the length of a newsgroup record \citep{blake1998uci}; eight mid-size data Tennessee Student Teacher Achievement Ratio (STAR) data \citep{DVN/SIWH9F_2008}, communities and crime data \citep{blake1998uci}, Bike rental count data \cite{fanaee2014event}, CPU data \cite{romano2021pmlb}, red wine quality data \cite{cortez2009modeling}, music original data \cite{romano2021pmlb}, residential house price data \cite{rafiei2016novel} and Tecator data \cite{romano2021pmlb}; and three data sets with large $N>5000$: online news popularity \citep{fernandes2015proactive}, blog feedback data set \citep{buza2014feedback} and Satellite image data \citep{romano2021pmlb}. We select classification data sets whose ARIs are all above $0.2$, which include data MNIST, Madelon, DNase, Theorem, and Amphibians. Besides, we add six additional publicly available benchmark data sets for the clustering task:  Iris species data \cite{blake1998uci}, WDBC Breast Cancer Wisconsin \cite{blake1998uci}, Tetragonula bee species data \cite{franck2004nest}, Chemical composition of ceramic samples data \cite{blake1998uci} and AFLP data of Veronica plants \cite{martinez2004species}. For the dimension reduction task, we exclude difficult data sets whose clustering accuracy on the first two reduced dimensions is always below the accuracy ARI cutoff $0.2$. As we tend to reduce the dimensionality of given data sets, the original data is better to have a larger number of features, for which we set the cutoff at $P > 15$. Due to computational expense, we also remove data sets that are too large, using a threshold of $N> 2000$. 
\section{Results}
\label{sec:results}
\subsection{Overview of the Results}
We address key IML questions Q1-Q3 in two sections, one for supervised tasks (classification and regression, and one for unsupervised tasks (clustering and dimension reduction). For supervised tasks, we consider feature importance rankings as our primary interpretation, whereas for unsupervised tasks we consider clustering results, both in the full dimensional space and reduced dimension embeddings. In each category, we select and apply robust metrics to measure the reliability of the resulting interpretations. For each task, summary figures generated via the dashboard are used to assess each question. 
\begin{figure*}[h!]
\centering
\includegraphics[width=\linewidth]{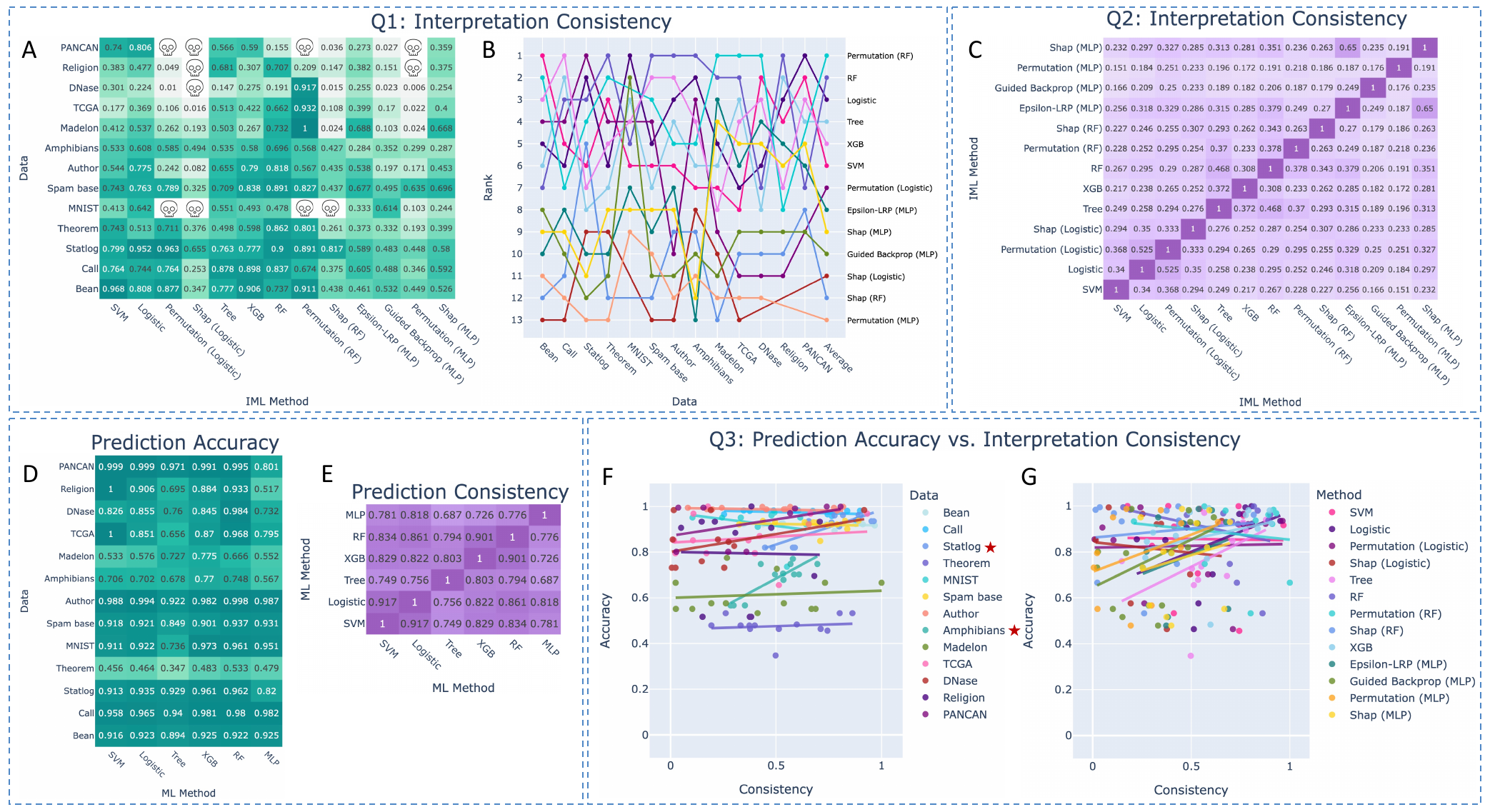}
\caption{\textbf{IML Performance on Classification Tasks.} \textbf{A:} Heatmap of within-method interpretation stability. \textbf{B:} Bump plot of IML methods ranked by the level of interpretation stability. \textbf{C:} Heatmap of between-method interpretation stability. \textbf{D:} Heatmap of between-method prediction accuracy on test sets. \textbf{E:} Heatmap of prediction stability on test sets. \textbf{F:} Scatterplot of accuracy and interpretation stability, colored by data sets, with fitted OLS lines aggregated over data. \textbf{G:} Scatterplot of accuracy and interpretation stability, colored by data sets, with fitted OLS lines aggregated over IML methods.}
\label{fig:f2}
\end{figure*}
For example, in Figure~\ref{fig:f2}, subplots (A) and (B) we aim to address Q1 by measuring whether interpretations are consistent among repeats, within each method. The heatmap (A) demonstrates the average within-method stability scores of interpretations of an IML method aggregated over 100 repeats, ranging in [0,1], with IML methods on the x-axis and data set on the y-axis. The bump plot (B) further addresses Q1 by ranking IML methods by their stability in each data set. The data sets are ordered by the \# observation/\# feature ratio, with the left y-axis showing the rank of each method based on stability. We add a column of average stability over all methods at the right, and the methods of the y-axis on the right are ordered by the average stability, from the most consistent to the least consistent. 

The heatmaps in (C) answer Q2 by evaluating whether different methods would result in similar interpretations on the same perturbed data via between-method average stability of interpretations obtained from each pair of IML methods. The heatmaps in panel (D) present each method's task-specific accuracy on the test set of each data, all averaging over 100 repeats. Heatmaps in (E) illustrate the average stability of prediction on the test set. 

To address Q3, we investigate the relationships between interpretation stability and prediction accuracy with scatterplots (F) and (G). Different colors represent different data sets in (F), and different colors represent different IML methods in (G), all averaging over 100 repeats. In each scatter plot, we also visualize the relationships by fitting regression lines, either aggregated over data or IML methods. Note that for the unsupervised tasks of clustering and dimension reduction, we cannot generate panels (E)-(G), due to the nature of the unsupervised tasks. Using these tools, we directly address Q1-Q3 for all supervised and unsupervised tasks. In Appendix \ref{app:results}, we assess the performance of specific methods for each question and task in more detail.






\subsection{Supervised Learning: Feature Importance (Classification \& Regression)}

\begin{figure*}[h]
\centering
\includegraphics[width=\linewidth]{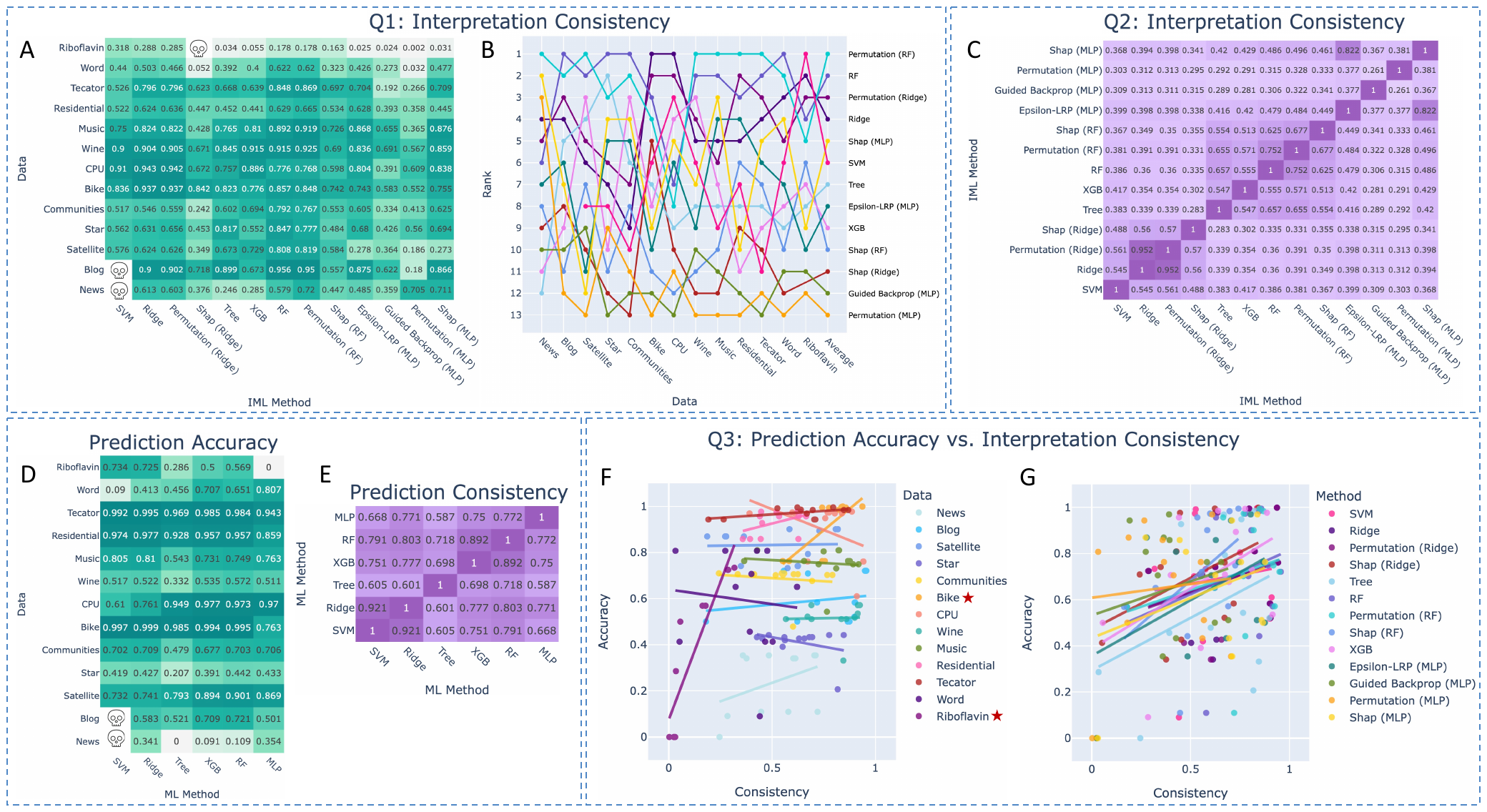}
\caption{\textbf{IML Performance on Regression Tasks.} \textbf{A:} Heatmap of within-method interpretation stability. \textbf{B:} Bump plot of IML methods ranked by the level of interpretation stability. \textbf{C:} Heatmap of between-method interpretation stability. \textbf{D:} Heatmap of between-method prediction accuracy on test sets. \textbf{E:} Heatmap of prediction stability on test sets. \textbf{F:} Scatterplot of accuracy and interpretation stability, colored by data sets, with fitted OLS lines aggregated over data. \textbf{G:} Scatterplot of accuracy and interpretation stability, colored by data sets, with fitted OLS lines aggregated over IML methods.}
\label{fig:f3}
\end{figure*}

\paragraph{Q1:}
Overall, the feature importance measures generated by IML methods are have low reliability with small stability scores, as shown in part (A) in Figures~\ref{fig:f2} and ~\ref{fig:f3}. For an individual data set, different IML methods also result in different stability levels in feature importance tasks. For example, in the DNase data, interpretations from permutation with RF are consistent with over AO of over 0.9, but the AO scores of other IML methods are all below 0.3. Comparing different tasks, the interpretations in regression are more consistent than those in classification with higher scores in part (A), using the same IML models. In panel (D) of Figures~\ref{fig:f2}, ~\ref{fig:f3}, the accuracy performance is consistent across methods. Specifically on feature importance tasks, by pairing stability and predictability analyses in Figure~\ref{fig:f2} and Figure~\ref{fig:f3}, methods have significantly different stability (panel (A)) even though they have similar predictive performance (panel (D)), such as in the PANCAN, Author, Statlog, Bean, and Call data sets in the classification task. Therefore, supervised models with similar accuracy may not have similar interpretation reliability.
\paragraph{Q2:}
Feature importance interpretations from different IML methods are not consistent with each other, with small stability scores for the between-method stability heatmaps in panel (C) of Figure~\ref{fig:f2} and Figure~\ref{fig:f3}. Tree based methods and linear models have the best between-method results, and permutation methods are more consistent with their base models. Combining with results panels (E), we can infer IML methods generate different feature importance scores on the same data, even with strong prediction strength.

\paragraph{Q3:}
To explore whether predictive accuracy can be used as an indicator for interpretation stability, we investigate their relationship by scatterplots in (F) and (G) of Figure~\ref{fig:f2} and Figure~\ref{fig:f3}. In panel (F) in both figures, most of the fitted lines are flat, and only two data sets have significant coefficients. Hence on a given data set, the higher predictive accuracy of an IML method does not imply higher interpretation stability. Note that in Riboflavin data in regression, MLP-related methods perform poorly in terms of both prediction accuracy and interpretation stability with almost 0 values, which causes a significant p-value between predictive accuracy and interpretation stability. From the perspective of IML methods, as shown in part (G), for a given method, higher accuracy on one data set does not imply higher interpretation stability. Though there are several methods that have positive coefficients, their coefficients are not significant, with p-values equal to 1 after Bonferroni correction due to the high variance of the results from different data sets. Therefore, none of the IML methods show significant associations between interpretation stability and predictive accuracy. 


The heatmap in (D) shows the average predictive accuracy on test sets of each ML method on each data, and the accuracy scores are similar across the ML methods. None of the methods have high accuracy scores in Madelon, Amphibians, and Theorem, which are more difficult to classify. And all of the methods have over 0.9 accuracies in Author, Call, and Bean data sets. Linear models and tree-based models result in high accuracy score of over 0.9 in most of the data sets, except the decision tree is less accurate in most of the high dimensional data. MLP is also relatively accurate in most of the data sets. 

\subsection{Unsupervised Learning: (Dimension Reduction \& Clustering)}
\begin{figure}[h!]
\centering
\includegraphics[width=1\linewidth]{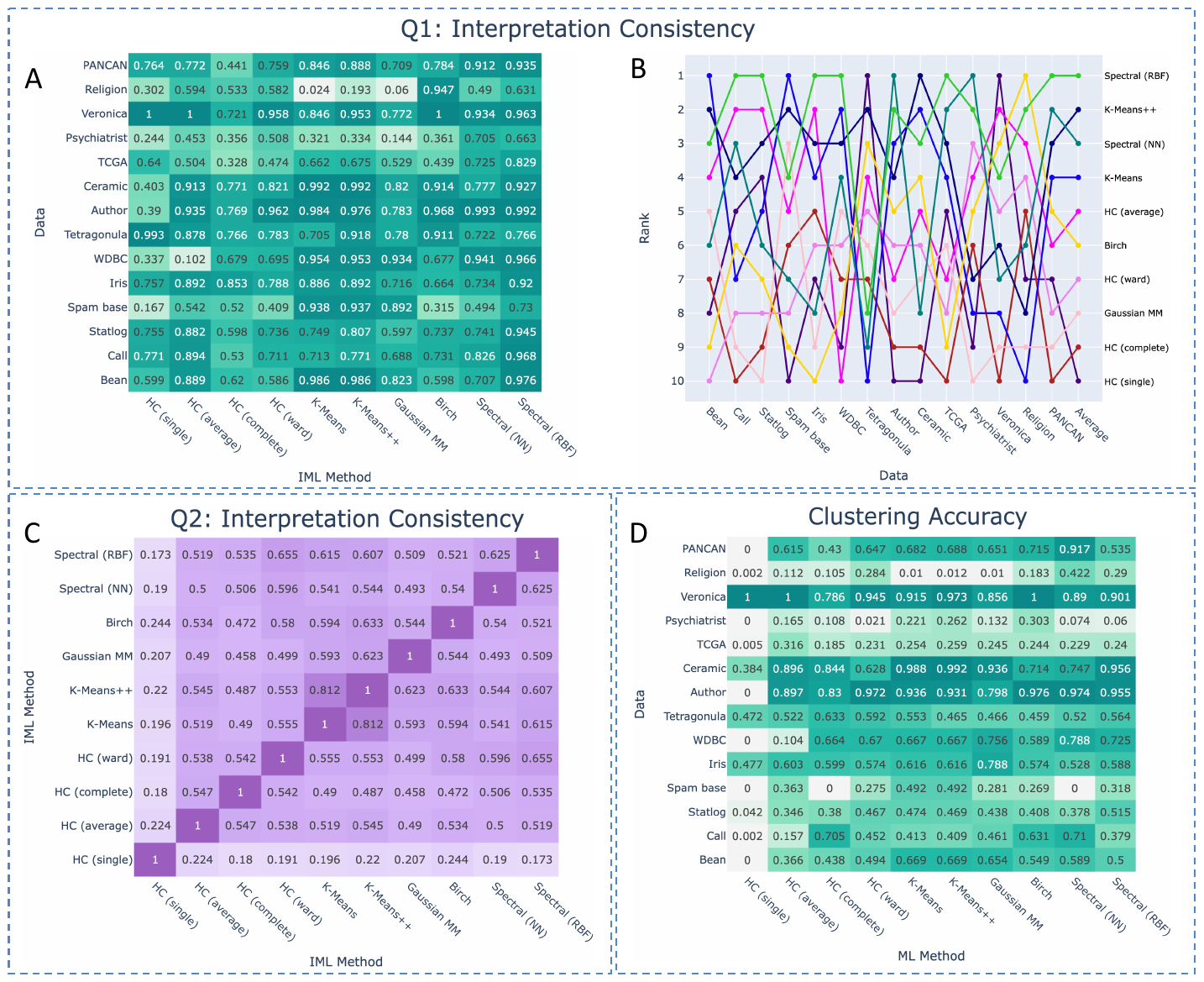}
\caption{\textbf{IML Performance on Clustering Methods}. \textbf{A:} Heatmap of within-method interpretation stability. \textbf{B:} Bump plot of IML methods ranked by the level of interpretation stability. \textbf{C:} Heatmap of between-method interpretation stability. \textbf{D:} Heatmap of between-method prediction accuracy.}
\label{fig:f4main}
\end{figure}

\paragraph{Q1:}
\paragraph{Clustering} In Figure~\ref{fig:f4main}, we can evaluate the IML tasks on the clustering results. In the bump plots of ranked feature importance in part (B),  we can observe that methods can have varying levels of stability in different data sets, regardless of the data sizes or prediction accuracy. Similar to the feature importance results in supervised tasks, we still find that on a single data set, different IML methods result in different stability levels. However, compared to classification and regression results which favored tree-methods and linear models, the overall stability scores in clustering are much higher. In the clustering task, on average, the spectral (RBF) and K-Means++ generate the most consistent clustering over all data sets, and HC (single) generates the least consistent clustering labels in data splitting.


\begin{figure}[h!]
\centering
\includegraphics[clip,width=1\textwidth]{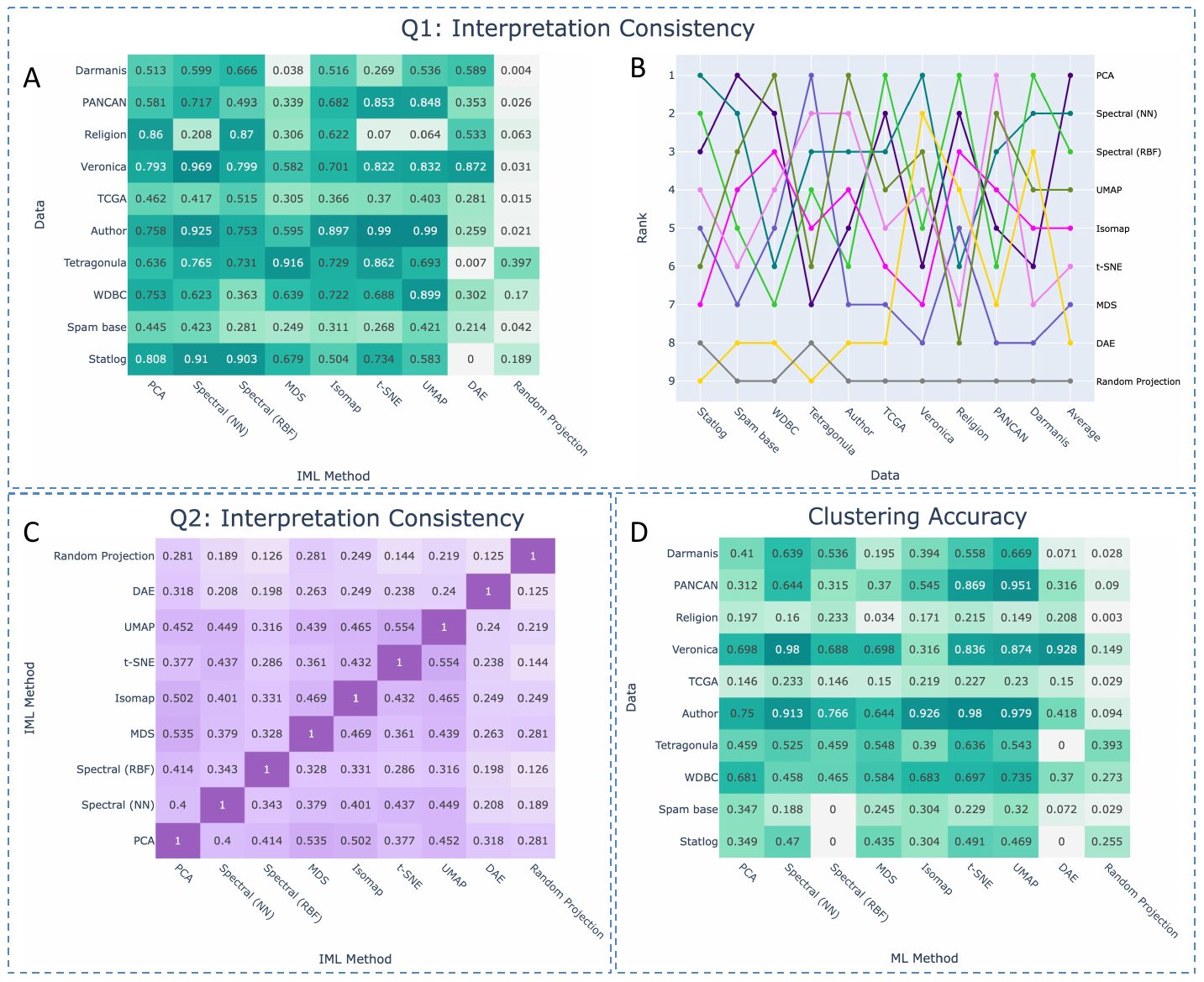}
\caption{\textbf{IML Performance on Dimension Reduction Methods}. \textbf{A:} Heatmap of within-method interpretation stability. \textbf{B:} Bump plot of IML methods ranked by the level of interpretation stability. \textbf{C:} Heatmap of between-method interpretation stability. \textbf{D:} Heatmap of between-method prediction accuracy. }
\label{fig:dr}
\end{figure}

\paragraph{Dimension Reduction} We evaluate the stability from two perspectives in the dimension reduction tasks: 1) stability on clustering performed in the reduced dimension and 2) the stability of an individual observations nearest neighbors. In Figure~\ref{fig:dr}(A) and (B), we can see the within-method clustering label stability. The stability is measured by the ARI metric. Plot (A) and (B) in Figure~\ref{fig:knn} analyze nearest neighbor-based stability with noise addition perturbation using random Gaussian noise with a standard deviation of 0.15, measured by the NN-Jaccard-AUC score, as explained in Section\ref{sec:metric}. From both perspectives, we reach similar answers to Q1, that the interpretations are not reliable within-method. 

In Figure~\ref{fig:dr}(D), we carry out clustering on the reduced dimension embedding as a notion of DR accuracy, with results showing similar inconsistency across methods and datasets. We utilize random projections here as a baseline model, which logiaclly yields the lowest accuracy and stability overall. The local methods t-SNE and UMAP work well for the bulk cell RNA-seq PANCAN data, but have poor performance in the single cell RNA-seq Darmanis data, possibly due to the sparsity of scRNA-seq data. In addition, inaccurate methods/data may have consistent clustering labels. For example, TCGA data has 0 clustering accuracy for all methods, but its labeling stability is moderate. The global methods PCA and spectral clustering can result in consistent clustering labels on average, as shown in plot (B) in Figure~\ref{fig:dr}. 

The nearest neighbor stability is relatively low for most methods and data sets, as shown in  plot (A) Figure~\ref{fig:knn}. Comparing the within-method stability heatmaps (A) in Figure~\ref{fig:dr} and Figure~\ref{fig:knn} and bump plots (B) in Figure~\ref{fig:dr} and Figure~\ref{fig:knn}, on the same reduced dimensions, the stability ranks of nearest neighbors are quite different from that of clustering. For example, even though the clustering labels based on DAE are unstable, the nearest neighbors are consistently maintained for larger values, especially for the Statlog and Tetragonula data sets. The bump plot Figure~\ref{fig:knn}(B) shows that DAE generates the most consistent nearest neighbors, while it is the least consistent in clustering.  

\paragraph{Q2:}
Between-method stability performance for clustering interpretations is moderate, with most of the ARI values around 0.5. K-Means and K-Means++ logically highly consistent with each other with an ARI of 0.812. All other methods are similarly stable, except HC (single), which has poor performance overall. 

We further consider the clustering label stability across different methods on the reduced dimension. The overall stability scores across different methods are moderate with most of the ARI around 0.5, as shown in the heatmap (C) in Figure~\ref{fig:dr}, similar to the clustering results. Clustering labels based on t-SNE are relatively consistent with PCA, Spectral (NN), and UMAP with ARI greater than 0.7, while PCA, Spectral (NN), and UMAP are not very consistent with each other. 

\paragraph{Q3:}
Note that under the task of clustering and dimension reduction, we treat the clustering labels our interpretation of the underlying structure of the data set. Therefore, Q3 does not apply here as the model accuracy is also calculated by the resulting labels.




\subsection{Dashboard and Software}

Since our empirical study covers a wide range of IML tasks, methods, benchmark data sets, perturbation techniques, and stability metrics, we are only able to show a small part of the analyses in the manuscript, and refer to Appendix \ref{app:results} for further results. Moreover, in order to demonstrate the full results with not only summary figures but also detailed figures, we created a user-friendly and interactive visualization dashboard\footnote{\url{https://iml-reliability.herokuapp.com/home}} based on the Python Dash developed by Plotly \cite{plotly}, with all data, methods, and metrics shown in this study. On this dashboard, researchers can also assess the reliability of their own data, IML method, or metric under this same framework. We have further developed an open-source Python package\footnote{\url{https://github.com/DataSlingers/IML_reliability}} to conduct the same framework of reliability tests, with functions that allow users to easily upload, evaluate, and compare their own reliability results. Further detailed description of software usage can be found in Appendix \ref{app:dashboard}.

\section{Discussion}

Through our extensive empirical study, our findings demonstrate that interpretations generated by commonly used interpretable machine learning (IML) methods are often unstable and sensitive to minor changes in the data. Across multiple tasks and datasets, we observe that small perturbations, such as train/test split variation or additive noise, can lead to substantial shifts in feature rankings, cluster assignments, or low-dimensional embeddings. This suggests that even when prediction accuracy remains unchanged, the interpretations provided by the model may not be trustworthy. Furthermore, different IML methods often produce divergent interpretations of the same data, and the consistency of a given method can vary significantly across datasets. Together, these results underscore that interpretation reliability is not a guaranteed byproduct of good predictive performance and must be assessed independently.

A central takeaway from our study is that no single IML method consistently yields the most stable or reliable interpretations across all datasets or tasks. The bump plots shown in panels (B) of Figures~\ref{fig:f2}, \ref{fig:f3}, \ref{fig:f4main}, and \ref{fig:dr} illustrate this clearly: different methods rank highly on different datasets, and their relative stability fluctuates depending on the task and data characteristics. For instance, in clustering tasks, the most stable method varies from dataset to dataset without a clear relationship to factors such as the number of features or the sample size. This aligns with the “no free lunch” principle in machine learning—there is no universally optimal method for interpretation reliability. Instead, method choice should be guided by empirical testing on the specific dataset at hand.

Although prior literature often advocates for simple, interpretable models like linear regression or shallow trees \cite{liu2020impact}, our findings challenge the assumption that such models are inherently more reliable in their interpretations. While these methods may be easier to understand, they do not always produce the most stable feature rankings or clustering outputs. In several instances, more complex or model-agnostic methods, such as permutation-based importance with random forests, yielded more consistent interpretation scores than their simpler counterparts. This introduces an important trade-off: simpler models may be more transparent, but that transparency does not always translate into stability. Critically, we also find that predictive accuracy is not a reliable proxy for interpretation stability. A highly accurate model may still yield volatile or misleading explanations. These results suggest that without explicit testing for stability, it is difficult to justify trusting the outputs of IML methods in sensitive decision-making settings. In practice, we recommend researchers to weigh the computational cost, interpretability, and stability of the methods they use.

While our study spans a diverse range of datasets and methods, it is limited to tabular data. Future work should extend the reliability framework to other data modalities such as images, text, and time series, where interpretation methods tend to differ and may be differently affected by data or model perturbations. Besides numerical interpretations, numerous formats of explanations can be considered in future studies, including rules, textual, or visual interpretations. Additionally, future work may be devoted specifically to inference methods to statistically quantify uncertainty in interpretation metrics. Recent advances in selective inference, post-selection inference, and model-agnostic feature attribution provide rigorous tools for estimating the variability and significance of interpretations \citep{berk2013valid, lee2016exact, tibshirani2016exact, mentch2016quantifying, gan2022inference, williamson2021general}. These methods can complement empirical stability analyses by offering confidence intervals or p-values for interpretations, enabling users to distinguish signal from noise in a statistically principled way.

Given our findings, we recommend for a new standard of practice in IML research and application: interpretation reliability should be validated and reported alongside predictive performance. Whenever possible, researchers and practitioners should evaluate their interpretations under stability measures defined here - perturbations such as resampling, noise addition, and random re-initializations, to assess robustness. This practice is especially important in contexts involving knowledge discovery from new datasets, development of novel IML algorithms, or proposals of new metrics for interpretability. Without such assessments, conclusions drawn from model interpretations may be overstated or misleading, leading to negative downstream impacts. Our dashboard and available Python package are valuable open-source, easy-to-use tools towards these aims. Ultimately, the reliability of interpretations affects not just academic analyses but also real-world decisions in healthcare, finance, and public policy. By advancing methods that ensure greater reliability in interpretation, we not only strengthen the foundation for scientific discovery, but also build the underlyign societal trust necessary for the responsible deployment of machine learning systems in these critical domains.

\paragraph{Acknowledgments} G.I.A acknowledges funding from NSF DMS-1554821. 
\newpage
\appendix
\section{Study Design}
\label{app:studydesign}

\subsection{Data Randomization}

We define an interpretation to be reliable if the same or similar interpretations can be derived from new data of the same distribution. Therefore, the reliability of a machine learning model can be measured by the stability of its derived interpretations, via sensitivity tests \citep{roscher2020explainable}, or randomly permuting data. In deep learning these permutations are exemplified through methods such as occlusion sensitivity tests \cite{zeiler2014visualizing} to analyze important parts of images by systematically occluding different portions of images, and saliency maps \cite{adebayo2018sanity} for random parameters. Interpretations for surrogate models such as LIME \cite{zhang2019should} investigate the uncertainties by its variance in explaining a single point and under different parameter choices, but such tests are highly model specific.  For unsupervised learning methods, \cite{wiwie2015comparing} apply noise addition to measure the robustness of clustering results, and other literature such as \cite{ziyan2009consistency} study the stability of their proposed clustering algorithm by randomly changing the initial data labels, or through bootstrapping for embeddings \cite{nguyen2019ten} or feature importance scores \cite{liu2020impact}.

A reliable machine learning model should not be overly sensitive to small changes in the data or parameters of the model. To this end, the first step of our framework is to design sensitivity tests to obtain interpretations from machine learning models under different perturbations. We propose two types of perturbation techniques to conduct sensitivity tests: 1) random splits and 2) noise addition. In supervised learning, researchers conduct a train/test split before fitting a predictive model so as to avoid overfitting. The interpretations are reliable if the resulting feature importance scores can remain unchanged and consistent across both 1) and 2). Therefore, in supervised models, we measure the stability of the top K important features ($K \in [1,30]$) obtained from 100 repeats of 70\%/30\% train/test splits. For unsupervised learning tasks including clustering and dimension reduction, we also implement sensitivity tests by subsampling 70\% of the data sets. Secondly, we add random noise during the unsupervised learning tasks of clustering and dimension reduction. The stability of interpretations and their resistance to additional noise can be obtained by adding random noise to the data of interest, and measuring the changes of results. With increasing levels of noise added, we are able to illustrate how the stability would change with more difficult data. Noise generated by either Normal ($N(0,\sigma^2)$)or Laplace ($Laplace(0,\sigma^2)$) distribution with mean zero is added to the original data set. The level of noise is controlled by variance $\sigma^2 \in [0,5]$, with higher variance indicating more noisy and difficult data. For dimension reduction methods, we additionally investigate the stability against the number of reduced dimensions(rank = 2, 5, 10).

\subsection{Stability Metrics}\label{app:secmetric}
\subsubsection{Interpretation Stability Metrics}
We include three categories of interpretability tasks: 1) feature importance/ranking derived from supervised learning, 2) clustering result, and 3) interpretations, namely clustering results, in reduced dimensions. Below we discuss metrics used to measure reliability of these interpretations in more detail.

\paragraph{Feature importance ranking}  Kendall's Tau distance \citep{kendall1938new}, which measures the total number of pairwise inversion, and Spearman footrule distance \citep{wackerly2014mathematical}, which is the L1 distance between ranks, are the two most popular metrics to compute rank similarities in the area of information retrieval. 
However, these metrics are indifferent to the top ranks and bottom ranks, while researchers are more interested in the stability of top rankings. \cite{shieh1998weighted} propose weighted Kendall's Tau statistics that can impose higher weights on items with top rankings. \cite{kumar2010generalized} propose weighted generalized versions of the two metrics by considering element weights, elements similarity, and position weights. \cite{yilmaz2008new} propose a new rank coefficient: AP correlation, which aims to put more weight on errors in top ranks. In terms of measurement of ranking on feature importance, \cite{liu2020impact} explore the impact of prediction accuracy on the quality interpretability in terms of accuracy and stability of feature importance ranking. \cite{liu2020impact} construct a set of features rankings through data bootstrapping utilize a weighted Kendall's Tau distance as the stability metric for feature ranking interpretation, by adding a $\frac{1}{k}$ penalty on bubble swap involving the $k^{th}$ rank. A smaller average pairwise distance indicates higher stability. \cite{liu2020impact} also propose to measure the interpretation accuracy by bounding the probability of true ranking that a given ranking is the ground truth by the probability that the ranking is equal to the mode. 

 As researchers are generally more interested in the most important features, we focus on top-K rank stability in the case of feature importance. One commonplace metric is the Jaccard similarity \citep{real1996probabilistic} of top-K features. With ranks $A$ and $B$, the Jaccard similarity is given by 
\begin{equation}
    J(A,B)@k =\frac{|A_k\cap B_k|}{|A_k\cup B_k|}. 
\end{equation}
where $A_k$ and $B_k$ contain only the top k features. 

In order to measure the top-k stability,  \cite{fagin2003comparing} provide $K^{(p)}$, the Kendall distance with penalty parameter $p$, which is determined by the co-occurrence of items within the top K ranks.  
\begin{equation}
    K(A,B)^{(p)}@k =\sum_{i,j \in \mathcal{P}(A_k,B_k)} \bar{K}_{i,j} ^{(p)}(A_k,B_k). 
\end{equation}

where $\mathcal{P}(A,B)$ is the set of all unordered pairs of elements in $A_k\cup B_k$ and $\bar{K}_{i,j} ^{(p)}(A_k,B_k)$ varies depending on whether $i$ and $j$'s co-occurrence in $A_k$ and $B_k$:  

Furthermore, \cite{fagin2003comparing} propose unbounded set-intersection overlap, which is also called average overlap (AO) \cite{webber2010similarity}. Specifically, AO is calculated as the average agreement of $A$ and $B$ of each depth and is given by
\begin{equation}
    AO@k  = \frac{1}{k} \sum_{d=1}^{k} \frac{|A_d\cap B_d|}{d}
\end{equation}
In this study, we implement these three widely used top K rank metrics including Jaccard similarity, top K Kendall's Tau, and AO to measure the top K feature rankings. The Jaccard similarity and AO range in $[0,1]$, and the top K Kendall's Tau range in $[-1,1]$, with a higher value indicating higher stability. The difference is that the Jaccard similarity and top K Kendall's Tau consider whether two sets contain the same elements, while the AO can further measure the stability of ranking.

\paragraph{Clustering}

The stability of a clustering method can be measured by the average pairwise similarity of clustering results under the same setting in the sensitivity test. To quantify the similarity between two clustering results: $A$ and $B$, a widely used metrics \textit{Adjusted Rand Index} (ARI) \citep{rand1971objective} represents the frequency of occurrence of agreements over the total pairs, with correction for chance, which is given by 
\begin{equation}
    ARI  = \frac{\sum_{ij} {n_{ij}\choose 2} - [\sum_{i} {a_{i}\choose 2} \sum_{j} {a_{j}\choose 2}]/ {n\choose 2}    }{\frac{1}{2}[\sum_{i} {a_{i}\choose 2} + \sum_{j} {a_{j}\choose 2}] - [\sum_{i} {a_{i}\choose 2} \sum_{j} {a_{j}\choose 2}]/ {n\choose 2}   }
\end{equation}

where $n_{ij}$ denotes the number of observations in common between cluster $i$ in A and cluster $j$ in B, and $a_i = \sum_{j\in B} n_{ij}$,  $b_i = \sum_{i\in A} n_{ij}$.   

We also utilize Fowlkes mallows index \citep{fowlkes1983method}, which is the geometric mean between the precision and recall: 

\begin{equation}
    FM = \sqrt{PPV \cdot TPR}
\end{equation}

where $PPV$ is the positive predictive rate and $TPR$ is the true positive rate. 

Another common clustering similarity metric is mutual information (MI) \citep{kraskov2004estimating}:
\begin{equation}
    MI(A,B)  = \sum_{i\in A}\sum_{j\in B} p(i,j)\log\frac{p(i,j)}{p(i)p(j)}
\end{equation}

where $i$ and $j$ represent clusters in A and B, respectively. The V measure \citep{rosenberg2007v} calculates the harmonic mean between homogeneity and completeness, given by 
\begin{equation}
     V = \frac{(1+\beta)*\text{homogeneity}*\text{completeness}}{\beta \text{completeness}+\text{homogeneity}}
\end{equation}

In this study, we use default $\beta=1$. Note that the V measure is equivalent to normalized mutual information. 

All of the metrics are ranged in $[0,1]$, with a higher value indicating higher stability. As evaluated in \cite{meilua2007comparing}, the adjusted rand index(ARI) and Fowlkes mallows index measure clustering similarity based on counting pairs of points in which two clustering results agree, while V measure and mutual information measure the variation of information. We implement these four metrics to have create a multifaceted perspective on stability. We further apply the average pairwise stability to measure the stability of the clustering techniques and obtain clustering accuracy by applying the same metrics to the predicted clustering labels and true labels.

We perform sensitivity tests on clustering methods using noise addition and data splitting with 70\% of the data. However, in the case of data splitting, since each subset contains different samples, we only compare the samples belonging to the intersection of every pair of splits and utilize the same clustering similarity metrics as noise addition. For example, if split 1 contains samples $[a,b,c,d,e]$, and split 2 contains $[a,c,d,e,f]$, we measure the similarity of clustering results of only samples $[a,c,d,e]$ using ARI, Fowlkes mallows, V measure and mutual information. 


\paragraph{Dimension reduction}

\begin{figure}[h!]
\centering
\includegraphics[clip,width=0.6\textwidth]{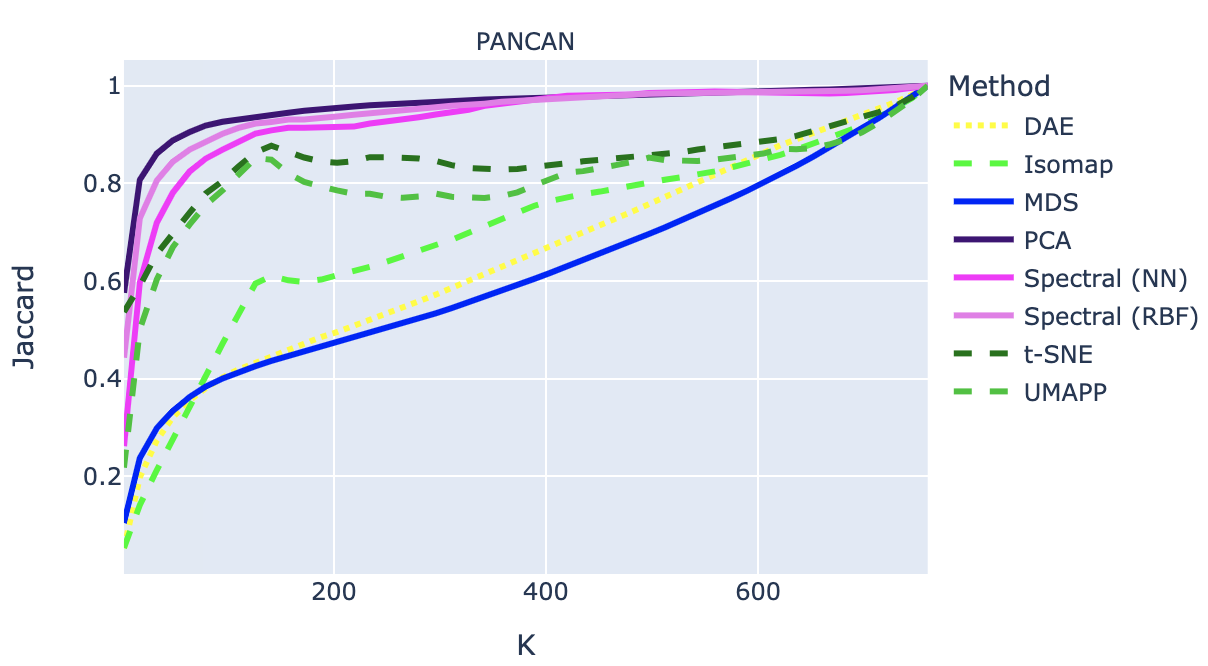}
    \caption{ROC curve of Jaccard score with the number of nearest neighbors K}
    \label{fig:line2}
\end{figure}

We measure the reliability of dimension reduction from two aspects: 1) visualization reliability measured by local neighbor stability and 2) clustering stability performed on the reduced dimension. Therefore, we develop a metric, denoted as \textit{NN-Jaccard-AUC} score, to measure the local neighbor stability in the reduced dimensions, for each sample, we can examine its nearest neighbors and check if this sample has a consistent set of nearest neighbors within different random replicates, under the same setting. Specifically, the similarity of two sets of nearest neighbors can be calculated by the Jaccard score \citep{real1996probabilistic}. The number of nearest neighbors to measure can be ranged from 1 to $N$, where $N$ is the total number of samples. Specifically, for $K = N$, the stability measured by the Jaccard score is always $1$ as the N-nearest neighbor contains the whole set of observations. We obtain the overall Jaccard scores by averaging the score for every sample. We then use the curve of Jaccard scores computed from $K = 1$ to $K=N$ against $K$ as a receiver operating characteristic curve, and we can further obtain the area under the curve ($AUC$) of the Jaccard scores curve. Hence, the AUC score of the Jaccard curve provides a quantitative stability metric on local similarity. A higher value of $AUC$ indicates higher similarities between two reduced dimensions, where $AUC  = 1$ indicates that the dimension reduction method is perfectly consistent in terms of local similarity under any $K$ range. Empirically, we construct the average Jaccard score with $50$ values of K, ranging from 1 to $N$, and we average the Jaccard score of $500$ randomly sampled samples under each $K$ to save computational cost when the number of samples is too large ($> 500$). The ROC curve of the jaccard score for dimension reduction is shown in an example dataset PANCAN in Figure~\ref{fig:line2}. One may ask why not use the AO or Kendall's Tau, which are used as metrics in the feature ranking stability. However, as we increase the number of neighbors $K$, the values would be dominated by the top neighbors so that the results would be less informative with large $K$.

On the other hand, using the dimension reduction techniques as a pre-processing procedure, we can measure the stability of downstream clustering performed on the reduced dimensions. Therefore, we apply clustering algorithms including both hierarchical and K-Means clustering to the reduced dimensions under the setting of the oracle number of clusters. Following the same framework in the clustering stability, we measure the stability of clustering results on the reduced dimension by the pairwise ARI, MI, V\_measure, and Fowlkes mallows index. All of the metrics range from $[0,1]$, with large values indicating stronger stability.



    


\subsubsection{Prediction Stability in Supervised Models}

If the training models are similar under different training sets, would they also generate similar interpretations? This question is specific to supervised learning methods, where we wish to explore the relationship between the stability of predicted estimates and the stability of feature importance interpretations. Similar to interpretation stability, we measure the prediction stability within each method as well as across different methods. 

\paragraph{Within-Method Prediction Stability} Specifically, for each split, we build a prediction model using the training set, record the feature importance scores, and make predictions on the test set. Then we measure the prediction stability of each sample by leveraging the dissimilarity of its predicted values.  However, since each train/test split contains different samples in the test sets, we are not able to directly compare test predictions from two splits. Instead, we evaluate the stability for each sample over 100 repeats of data splitting. In the classification task, the dissimilarity of prediction on a specific sample is evaluated by the entropy of its predicted classification groups, where the entropy of sample $i$ is defined as 
$$
entropy_i = - \sum_{k=1}^{K} p_k \log(p_k) 
$$
where $K$ is the number of classes and $p_k$ is the proportion of class $k$. In the regression task, the standard deviation of predicted values is used as the stability metric. In both scenarios, a smaller value indicates higher purity of the predicted responses, and we construct the stability scores by exponential negative dissimilarity values. 

\paragraph{Between-method prediction stability} On each train/test split, the similarity between test predictions generated from two IML methods can be evaluated as they are based on the same set of samples. In classification, we evaluate pairwise prediction stability by the proportion of the same predicted labels. And in regression, we compute the MSE between two test predictions and obtain stability scores by 1 minus the average of the min-max normalized MSE over all repeats.

\subsection{Interpretable Machine Learning Methods}
 We focus on the most popular machine learning and deep learning methods in these areas, developed reliable metrics to measure the reliability of the interpretability, and then conduct an empirical analysis to a wide range of real tabular data sets. Since we also aim to test the interpretations derived from generic global methods against each other, we do not apply posthoc methods which are specifically applied to certain machine learning models for convolutional neural network, or methods that provide local feature importance, such as LIME \citep{ribeiro2016should} or anchor \citep{ribeiro2018anchors}. Specifically, we include general model-specific machine learning models, deep learning related methods, and model-agnostic methods.

\subsubsection{Feature importance }
We apply the most popular general supervised models including linear models LASSO/Logistic LASSO \citep{tibshirani1996regression}, Ridge/Logistic Ridge \citep{tolles2016logistic}, tree-based models random forest \citep{breiman2001random}, XGradient Boosting (XGB) \citep{chen2015xgboost}, and linear SVM \citep{cortes1995support} in the case of classification. The optimal hyper-parameters are chosen by cross-validation. In addition, even though neural networks are black box models with no inherent interpretation abilities, there are plenty of post-hoc methods to generate feature importance by gradient-based attribution  \citep{selvaraju2016grad,shrikumar2017learning,simonyan2013deep,zeiler2014visualizing,sundararajan2017axiomatic,kapishnikov2019xrai,selvaraju2016grad}, layer-wise relevance propagation \citep{montavon2019layer,iwana2019explaining,gu2018understanding}, proxy model \citep{zilke2016deepred}, or knockoffs \citep{lu2018deeppink}. Therefore, we measure the stability of interpretations from neural network-based methods by fitting a multi-layer perception (MLP) model with two hidden layers, and internal ReLu activations with a softmax activation in the final layer. We extract feature importance scores from the MLP model by implementing a number of these post-hoc methods. We note that that we stopped any method that runs over 12 hours to finish a single repeat, such as permutation and Shapley value methods in high-dimensional data sets. 
\paragraph{Linear models}
\begin{itemize}
    \item {\textbf{LASSO \& Ridge}} Linear models are favored in applications with their straightforward interpretations. We evaluate the feature importance rankings by the magnitude of the regression coefficients, which reflects the change of response with respect to the unit change of the feature. The formula for LASSO regression is defined as 
$$
Y = X\beta +\lambda|\beta|, 
$$
and the formula for ridge regression is defined as 
$$
Y = X\beta +\lambda|\beta|_2^2, 
$$

where $\lambda$ is the level of penalty. In our implementation, we select a penalty via 5-fold cross-validation. 

    \item {\textbf{Linear SVM}} Support vector machine make predictions by creating a hyperplane that can maximize the distance between support vectors. Similar to least square models, we use the magnitude of the coefficients for feature importance ranking. 
\end{itemize}
\textbf{Implementation details: }
\begin{itemize}
    \item LASSO: sklearn.linear\_model.LassoCV function with L1 penalty is used to build the logistic lasso model. The penalty parameter is selected by cross-validation with 5 folds, saga solver, max number of iteration =100, and all other parameters are set as default. 
    
    \item Ridge: sklearn.linear\_model.RidgeCV function with L2 penalty is used to build the logistic lasso model. The penalty parameter is selected by cross-validation with 5 folds, saga solver, and max number of iterations =100, and all other parameters are set as default. 

    \item SVM (Regression): sklearn.svm.LinearSVR with L2 penalty is used used to build the SVM model. All parameters are set as default. 
    \item Logistic LASSO: sklearn.linear\_model.LogisticRegressionCV function with L1 penalty is used to build the logistic lasso model. The penalty parameter is selected by cross-validation with 5 folds, saga solver, max number of iteration =100, and all other parameters are set as default.

    \item Logistic Ridge: sklearn.linear\_model.LogisticRegressionCV function with L2 penalty is used to build the logistic lasso model. The penalty parameter is selected by cross-validation with 5 folds, saga solver, and max number of iterations =100, and all other parameters are set as default. 
    
    \item SVM (Classification): sklearn.svm.LinearSVC with L2 penalty is used used to build the SVM model. All parameters are set as default. 

    \end{itemize}

\paragraph{Tree-based methods} Tree-based methods obtain feature importance by the accumulated impurity decrease within each tree. The feature importance of feature $j$ can be calculated as 
$$
I_j = \frac{1}{B}\sum_{b=1}^B \sum_{s\in S_b} I(v(s)=j)(L_{s-1} - L{s}) 
$$

where $B$ is the number of decision trees, $S_b$ includes all nodes in tree $b$, $v(s)$ denotes the feature used to split node $s$, and $L_s$ denotes loss after splitting node $s$. Features with higher importance scores are able to generate more homogeneous sub-nodes. We implemented widely used tree-based models, including decision trees, random forests, and eXtreme Gradient Boosting (XGB).

\textbf{Implementation details: }
\begin{itemize}
    \item Decision tree (Classification): sklearn.tree.DecisionTreeClassifier function is used to build the decision tree model. All parameters are set as default. 
    
    \item Random forest (Classification): sklearn.ensemble.RandomForestClassifier function is used to build the random forest model. All parameters are set as default. 
        
    \item XGB (Classification):  xgboost.XGBClassifier is used used to build the XGB model. The "objective" paramter is set to be "multi:softmax" if the number of classes is greater than 2. And all other parameters being defaulted. 
         \item Decision tree (Regression): sklearn.tree.DecisionTreeRegressor function is used to build the decision tree model. All parameters are set as default. 
        
        \item Random forest (Regression): sklearn.ensemble.RandomForestRegressor function is used to build the random forest model. All parameters are set as default.

        \item XGB (Regression): xgboost.XGBRegressor is used to build the XGB model. All parameters being default.  
    
    \end{itemize}
\paragraph{Post-hoc deep learning methods. }
Despite the lack of interpretability in the deep neural networks models, post-hoc methods have been proposed to measure feature importance by assigning attribution value to the input features. 
Gradient-based methods compute attributions of features in a forward and backward path of the network \citep{adebayo2018sanity}, by applying chain rule of gradients. Popular methods include integrated gradients\citep{sundararajan2017axiomatic}, guided backpropagation \citep{springenberg2014striving}, saliency map \citep{simonyan2013deep}, with further work establishing DeepLIFT and Episilon-LRP as modified gradient backpropagation methods\cite{ancona2017towards, shrikumar2017learning, bach2015pixel}. Another set of methods is perturbation-based methods, which measure the marginal effect of the input features on the output neurons by replacing features with zero baselines \citep{zeiler2014visualizing}.

\begin{itemize}
    \item {\textbf{Saliency map }} Saliency map \citep{simonyan2013deep} constructs feature importance by partial gradients of input feature on the output target. 

\item {\textbf{Guided backpropagation}} Guided backpropagation \citep{springenberg2014striving} conduct backpropagation through ReLu units, by setting negative gradients to zeros. 

\item {\textbf{Integrated Gradients}} Integrated gradients\citep{sundararajan2017axiomatic} computes the average gradients by summing over various inputs from zero baselines to actual values sample. 

\item {\textbf{Epsilon-LRP}} Epsilon-LRP\citep{bach2015pixel} computes a layer's relevance to the target neuron, and recursively redistribution a layer's relevance until the input layer. 

\item {\textbf{deepLIFT}} Similar to Epsilon-LRP, DeepLIFT \citep{shrikumar2017learning} assigns input features' attributions by their relative effect on the output compared to some reference input.   

\item {\textbf{Occlusion}} \citep{zeiler2014visualizing} measures the marginal effect of features on the target output neurons by replacing input array with zeros on a rolling base. 

\end{itemize}
\textbf{Implementation:} 
\begin{itemize}
\item Regression: We build a multi-layer perception model with 2 hidden layers, internal ReLu activations, mean squared error loss, and adam optimizer. The MLP model is fitted with 10 epochs and a batch size of 50. 
    
        \begin{itemize}
        \item deepLIFT: the DeepLift package \cite{githubGitHubKundajelabdeeplift} with nonlinear\_mxts\_mode=NonlinearMxtsMode.RevealCancel (DeepLIFT-RevealCancel at all layers) to construct feature importance scores. We set \\ find\_scores\_layer\_idx=0, target\_layer\_idx=-1. The test data set is passed to the scoring function with batch\_size=100 and task\_idx = 0.

        \item Guided backpropagation masked: the DeepLift package with \\ nonlinear\_mxts\_mode=NonlinearMxtsMode.GuidedBackprop to construct feature importance scores, and we mask out positions that are zero.  We set find\_scores\_layer\_idx=0, target\_layer\_idx=-1. The test data set is passed to the scoring function with batch\_size=100 and task\_idx = 0.

        \item Epsilon-LRP: the DeepExplain package \cite{githubGitHubMarcoanconaDeepExplain} with "elrp" as method\_name is used to construct  Epsilon-LRP results. We set the first layer of MLP as the input layer and the last layer as the output.

        \item Integrated Gradients: the DeepExplain package with "intgrad" as method\_name is used to construct integrated gradient results. We set the first layer of MLP as the input layer and the last layer as the output.  
  
        \item Saliency map: the DeepExplain package with "saliency" as method\_name is used to construct saliency map results. We set the first layer of MLP as the input layer and the last layer as the output. 
              \item Occlusion: the DeepExplain package with "occlusion" as method\_name is used to construct occlusion feature importance. We set the first layer of MLP as the input layer and the last layer as the output.  
    \end{itemize}
    
    \item Classification: We first build a multi-layer perception model with 2 hidden layers, and internal ReLu activations with a softmax activation in the final layer, categorical cross-entropy loss, adam optimizer, and accuracy as the metric. The MLP model is fitted with 2 epochs and a batch size of 10. We use the pre-softmax layer to construct feature importance measures.  
        \begin{itemize}
        \item deepLIFT: the DeepLift package with nonlinear\_mxts\_mode=NonlinearMxtsMode.RevealCancel (DeepLIFT-RevealCancel at all layers) to construct feature importance scores. We set \\ find\_scores\_layer\_idx=0, target\_layer\_idx=-2. The test data set is passed to the scoring function with batch\_size=100 and task\_idx = 0.

        \item Guided backpropagation masked: the DeepLift package with \\ nonlinear\_mxts\_mode=NonlinearMxtsMode.GuidedBackprop to construct feature importance scores, and we mask out positions that are zero.  We set find\_scores\_layer\_idx=0, target\_layer\_idx=-2. The test data set is passed to the scoring function with batch\_size=100 and task\_idx = 0.

        \item Epsilon-LRP: the DeepExplain package with "elrp" as method\_name is used to construct  Epsilon-LRP results. We set first layer of MLP as the input layer and the pre-softmax layer as the output. For data with the number of classes greater than 2, we use the average scores across different classes as the global feature importance score.

        \item Integrated Gradients: the DeepExplain package with "intgrad" as method\_name is used to construct integrated gradient results. We set the first layer of MLP as the input layer and the pre-softmax layer as the output. For data with the number of classes greater than 2, we use the average scores across different classes as the global feature importance score. 
  
        \item Saliency map: the DeepExplain package with "saliency" as method\_name is used to construct saliency map results. We set the first layer of MLP as the input layer and the pre-softmax layer as the output. For data with the number of classes greater than 2, we use the average scores across different classes as the global feature importance score. 
              \item Occlusion: the DeepExplain package with "occlusion" as method\_name is used to construct occlusion feature importance.We set the first layer of MLP as the input layer and the pre-softmax layer as the output. For data with the number of classes greater than 2, we use the average scores across different classes as the global feature importance score.
    \end{itemize}

\end{itemize}
\paragraph{Model agnostic methods (applied to Logistic ridge, random forest, XGB and MLP)}

Model agnostic methods are post-hoc interpretable methods that can be applied to a wide range of prediction models. For the local model agnostic methods, we approximate the global feature importance by averaging the local feature importance score over all instances. 
\begin{itemize}
    \item {\bf{Permutation feature importance} } Permutation feature importance \citep{strobl2008conditional,strobl2007bias} is obtained by comparing the prediction errors after randomly permuting the feature of interest. It provides global insights into the effect of a feature on the prediction results. 

\item {\bf{SHAP}} Shapley values \citep{lundberg2017unified} provides local feature importance by computing the average marginal effect of a feature by considering all possible coalitions. The computation cost increases exponentially with the number of features. SHAP (Shapley Additive Explanations) provides additive explanations of features.  


\end{itemize}

\textbf{Implementations: }
\begin{itemize}

        \item Permutation
         \begin{itemize}
        \item Permutation Importance (Ridge/Logistic Ridge): the sklearn.inspection.permutation\_importance function is used to calculate permutation feature importance for the constructed ridge/logistic ridge model. The number of repeats is 10 and all other parameters are set as default. 
            
        \item Permutation Importance (RF): the sklearn.inspection.permutation\_importance function is used to calculate permutation feature importance for the constructed random forest model. The number of repeats is 10 and all other parameters are set as default.

        \item Permutation Importance (XGB): the sklearn.inspection.permutation\_importance function is used to calculate permutation feature importance for the constructed XGB model. The number of repeats is 10 and all other parameters are set as default.

            \item Permutation Importance (MLP): the eli5.sklearn.PermutationImportance function is used to calculate permutation feature importance for the MLP model. And all parameters are set as default. 

        \end{itemize}

        \item Shapley value \citep{lundberg2017unified}
        
        \begin{itemize}
        \item Shapley value (Ridge/Logistic Ridge): the shap.PermutationExplainer function is used to calculate shapley value for the constructed ridge/logistic ridge model. All parameters are set as default.
        \item Shapley value (RF): the shap.TreeExplainer function is used to calculate Shapley value for the constructed random forest model. All parameters are set as default.
        \item Shapley value (XGB): the shap.TreeExplainer function is used to calculate Shapley value for the constructed XGB model. All parameters are set as default.
            \item Shapley value (MLP): the DeepExplain package with "shapley\_sampling" as method\_name is used to construct Shapley value for the MLP model. In classification, we set the first layer of MLP as the input layer and the pre-softmax layer as the output. For data with the number of classes greater than 2, we use the average scores across different classes as the global feature importance score. In regression, we set the first layer of MLP as the input layer and the last layer as the output. 
        \end{itemize}
        

\end{itemize}









\subsubsection{Clustering}
To measure the stability of clustering interpretations, we apply the existing popular clustering techniques including K-Means based methods \citep{macqueen1967some,sculley2010web,arthur2006k}, hierarchical clustering with different linkage and euclidean distance 
\citep{murtagh1983survey}, spectral clustering with different distance affinity \citep{shi2000normalized}, Gaussian mixture model \citep{bishop2006pattern} and BIRCH \citep{zhang1996birch}, which is more efficient with large data sets. For all methods, the number of clusters is set to be the oracle number. 
\begin{itemize}
    \item {\bf{Hierarchical clustering}}  Hierarchical clustering seeks to build a nested dendrogram of the objects. Hierarchical clustering is quite flexible such that objects can be merged with different distance function and linkages, based on the nature of data set. Agglomerative Hierarchical Clustering \citep{murtagh1983survey} constructs the dendrogram using a bottom-up algorithm, where each object forms its own cluster, and similar clusters are merged together.  Each variation of hierarchical clustering uses different linkages:

\begin{itemize}
        \item Average linkage: The average linkage tend to merge two clusters with the minimum average distance between all observations in the two clusters. 
        

        \item Ward.D linkage: The Ward.D linkage tend to merge two clusters with the minimum sum of squared differences between observations in the two clusters. 
        
           \item Complete linkage: The Complete linkage tend to merge two clusters with the minimum distance between the farthest observations in the two clusters.

             
        \item Single linkage: The Single linkage tend to merge two clusters with the minimum distance between the closest observations in the two clusters.
        
    \end{itemize}
    
    For each data set, we select distance function with the highest clustering accuracy for the single, complete and average linkages, respectively, as specified in Table~\ref{table:dis}. We apply euclidean distance for Ward.D linkage. Additionally, the BIRCH algorithm (balanced iterative reducing and clustering) \citep{zhang1996birch} stores summary information using a dynamic clustering feature tree structure. And it is advantageous for large data sets and data with outliers \citep{xu2005survey}. 
    




    \item {\bf{ BIRCH}}:  The BIRCH algorithm (Balanced Iterative Reducing and Clustering) stores summary information about candidate clusters in a dynamic tree data structure. This tree hierarchically organizes the clusters represented at the leaf nodes. The tree can be rebuilt when a threshold specifying cluster size is updated manually, or when memory constraints force a change in this threshold. This algorithm has a time complexity linear in the number of instances.

 \item {\bf{K-Means \& extensions}} K-means \citep{macqueen1967some} partitions observations into clusters by minimizing the within-cluster euclidean distance. Starting with initial random centroids, the K-means algorithm iteratively clusters objects and constructs new centroids until convergence, with the goal of minimizing the within-cluster sum of squares. Numerous extensions to the K-means algorithm have been proposed, such as the Mini Batch K-Means \citep{sculley2010web} randomly selects a subset of observations at each iteration, so as to provide faster computation but lower accuracy than K-means. On the other hand, instead of starting with random initialization of centroids, K-means++ \citep{arthur2006k} initialize distant centroids, which may lead to higher accuracy than regular K-means.

  \item {\bf{Spectral clustering}}
   Spectral clustering \cite{von2007tutorial} first projects the affinity matrix of data into lower-dimensional embedding, then performs clustering methods on the resulting space. The affinity matrix can be constructed using a kernel function, such as radial basis function (RBF) kernel, or by computing a graph of nearest neighbors. 
   
   \item {\bf{Gaussian mixture model}} The Gaussian mixture model \cite{reynolds2009gaussian} assumes the observations are generated from a mixture of Gaussian distribution. It can be regarded as a generalized K-means algorithm but it takes the covariance structure of data into account.

\end{itemize}

\textbf{Implementations}
\begin{itemize}
  \item Hierarchical clustering + Average linkage: The sklearn.cluster.AgglomerativeClustering function with linkage "average" is used to construct hierarchical clustering. All parameters are set as default. 

        \item Hierarchical clustering + Ward.D linkage: The sklearn.cluster.AgglomerativeClustering function with linkage "ward" is used to construct hierarchical clustering. All parameters are set as default. 
        
           \item Hierarchical clustering + Complete linkage: The sklearn.cluster.AgglomerativeClustering function with linkage "complete" is used to construct hierarchical clustering. All parameters are set as default.
             
        \item Hierarchical clustering + Single linkage: The sklearn.cluster.AgglomerativeClustering function with linkage "single" is used to construct hierarchical clustering. All parameters are set as default. 
        
        \item BIRCH: The sklearn.cluster.Birch is used to construct Birch clustering. All parameters are set as default. 
        
        \item K-Means: The sklearn.cluster.KMeans is used to construct KMeans clustering. All parameters are set as default. 
         \item K-means++: The sklearn.cluster.KMeans is used to construct KMeans clustering. "init" is set as "k-means++" and all other parameters are set as default. 
       
        \item Mini Batch K-Means:  The sklearn.cluster.MiniBatchKMeans is used to construct MiniBatch K-Means clustering. All parameters are set as default, where the subsample size is set as  3 * n\_clusters. 
              \item Spectral clustering + nearest neighbors as affinity: The sklearn.cluster.SpectralClustering is used to construct Spectral clustering. "affinity" is set as "nearest\_neighbors" and all other parameters are set as default, where K-means clustering is used. 
        \item Spectral clustering + radial basis function (RBF) kernel as affinity: The sklearn.cluster.SpectralClustering is used to construct KMeans clustering. "affinity" is set as "rbf" and all other parameters are set as default, where K-means clustering is used. 
        
        \item Gaussian mixture model: sklearn.cluster.GaussianMixture is used to construct Gaussian mixture clustering. All parameters are set as default.

\end{itemize}

\subsubsection{Dimension reduction}

For dimension reduction methods, we select the most widely used linear global technique: principal component analysis (PCA) \cite{abdi2010principal}, and include random projection as a base line. For non-linear manifold learning methods, we select methods that preserve global properties such as metric/non-metric multidimensional scaling (MDS) \cite{davison2000multidimensional, agarwal2007generalized} and Isomap \cite{balasubramanian2002isomap}, and methods that preserve local properties including t-distributed Stochastic Neighbor Embedding (tSNE) \citep{van2008visualizing}, Uniform Manifold Approximation and Projection for Dimension Reduction (UMAP) \citep{mcinnes2018umap}, spectral embedding with different affinity \citep{shi2000normalized}. In addition, we apply a deep auto-encoder framework \citep{kramer1991nonlinear} with a three-layered encoder and a three-layered decoder, and the low-dimensional code produced by the encoder is used as the reduced dimensions.

   \paragraph{Linear models}
\begin{itemize}
\item \textbf{{PCA}} Principal component analysis linear transforms data of high dimensions to lower coordinates which can explain most of the variance of the original data.  
 
\item {\textbf{Gaussian random projection}} Random projection is a linear dimension reduction technique by applying multiplication to a random Gaussian matrix. The reduced dimensions can preserve the distances among points via Johnson-Lindenstrauss Lemma, and the algorithm is fast and robust to outliers. 
 \end{itemize}
\paragraph{Non-linear manifold learning}
\begin{itemize}
    \item \textbf{Global methods}
        \begin{itemize}
            \item {\textbf{Metric MDS(euclidean)}}  Multidimensional scaling generates lower dimension from given e pairwise distance between data points.
         
         \item {\textbf{Non-metric MDS}}  Non-metric MDS is a variant of MDS method which aims to preserve the rank of similarities. 
            
        \item {\textbf{Isomap}} Isomap is a nonlinear manifold learning method that conducts isometric mapping, using the geodesic distance from embedded neighborhood graph.
        \end{itemize}
 \item \textbf{Local methods}
    \begin{itemize}
            \item {\textbf{tSNE}} t-distributed stochastic neighbor embedding \citep{van2008visualizing} maps high dimensional data into 2 or 3 dimension. It first constructs a probability distribution that reflects similarities of points and then aims to minimize its KL divergence to another similar probability distribution in lower dimensions. t-SNE is popular for visualization purposes. 
 \item {\textbf{UMAP}} Uniform Manifold Approximation and Projection (UMAP) is another nonlinear dimension reduction technique that is popular for visualization. It aims to find lower dimensional data with the closest fuzzy topological structure as the original data.  
   \item {\textbf{Spectral embedding}} Spectral embedding utilizes the spectral of an affinity matrix of data and projects into lower-dimensional embedding.

\end{itemize}
\item \textbf{Deep Learning}
\begin{itemize}
    \item {\textbf{Deep Autoencoder}} Autoencoder is a type of neural network that encodes the information of data into a small number of neurons and then reconstruct the message by decoding. The code layer of the autoencoder can be regarded as the reduced dimensions of the input data.
\end{itemize} 
\end{itemize}

\textbf{Implementation}
\begin{itemize}
    \item PCA: The sklearn.decomposition.PCA is used to construct principal component analysis. All parameters are set as default.  
    \item Random projection: The sklearn.decomposition.GaussianRandomProjection is used to construct principal component analysis. n\_components is set as the given rank, and all other parameters are set as default.  
           \item {Metric MDS(euclidean)}: The sklearn.decomposition.MDS is used to construct MDS. n\_components is set as the given rank, metric as True, and all other parameters are set as default.  

        \item {non-Metric MDS}: The sklearn.decomposition.MDS is used to construct non-metric MDS. n\_components is set as the given rank, metric as False, and all other parameters are set as default.  
            
          \item {Isomap}: The sklearn.decomposition.Isomap is used to construct Isomap. n\_components is set as the given rank, and all other parameters are set as default.  
          
          \item tSNE: The sklearn.decomposition.TSNE is used to construct tSNE. n\_components is set as the given rank, learning\_rate as 100, and all other parameters are set as default.  
  \item UMAP: We set the learning rate as 1, number of maximum iteration as 200,  number of neighbors as 15 and minimum distance as 0.1
  \item Spectral embedding + nearest neighbors affinity: The sklearn.decomposition.SpectralEmbedding is used to construct Spectral embedding. n\_components is set as the given rank, "affinity" as "nearest\_neighbors, and all other parameters are set as default.  
        \item Spectral embedding + radial basis function (RBF) kernel: The sklearn.decomposition.SpectralEmbedding is used to construct Spectral embedding. n\_components is set as the given rank, "affinity" as "rbf", and all other parameters are set as default.  \item  Deep Autoencoder: We build an auto-encoder using 3 hidden layers in the encoder and 3 hidden layers in the decoder. The numbers of nodes in the hidden layers of the encoder are $M/6$, $M/12$, and $M/24$, respectively; the numbers of nodes in the hidden layers of the decoder are $M/24$, $M/12$ and $M/6$, respectively; and size of code is 2. We use internal ReLu activations with a sigmoid activation in the final layer, categorical cross-entropy loss, and adam optimizer. The Autoencoder model is fitted using 90\% of the data with 20 epochs and validated with the rest 10\%. 
\end{itemize}



 




\section{Additional Results}
This appendix provides detailed empirical results that support the main findings presented in Section~\ref{sec:results}. While the main text focuses on summarizing high-level insights, here we include extended figures and dataset-level observations across all tasks (classification, regression, clustering, and dimension reduction) to answer the reliability questions Q1--Q3.
\subsection{Q1: Within-method stability}

Figures~\ref{fig:f2}, \ref{fig:f3} summarize empirical results of feature importance methods in classification and regression tasks, measured with 10 top features and AO metrics. Note that the model-agnostic methods permutation and Shapley value are computationally inefficient and did not finish running one repeat within 12 hours in some of the large N or large P data sets.  Figure~\ref{fig:f4main} shows the summarized empirical results in the clustering task, measured with ARI metrics with data splitting, and Figure~\ref{fig:dr} summarized results for dimension reduction. 

Across tasks, the heatmaps of within-method stability (A) demonstrates overall, (1) the interpretations generated by IML methods are still not always consistent; (2) an IML method can have quite different levels of stability in different data sets, regardless of the data sizes; and (3) for a single data set, different IML methods result in different stability levels. The bump plots in (B) convey the message of no free lunch: there is no IML method that can work universally well in all data. The permutation (RF) and RF generate the most consistent interpretations over all data sets in both regression and classification.  Another takeaway from these figures is that simple methods are not necessarily the most consistent ones. Permutation (RF), which is a more complicated model, is quite consistent overall. But the trade-off is that it would take a longer time to compute. Shapely values and MLP-related methods are the least consistent ones on average. Combining with results in (A) and (C), we can infer that levels of interpretation stability of IML methods are quite different, even though they are similarly accurate, such as in the PANCAN, Author, Statlog, Bean, and Call data sets in classification. Therefore, models with similar accuracy may not have similar interpretations. 

\paragraph{Classification} In Figure~\ref{fig:f2}, we can see that the linear and tree-based IML models are more consistent in large N data sets, such as the strong performance in the high dimensional PANCAN data. MLP-based methods also work better in larger N data sets than high dimensional data, but they in general produce less consistent interpretation stability scores. The model-agnostic method permutation has good performance with linear or tree as base models. But another model-agnostic method Shapley value is not reliable in most of the data sets. 

We observe that a single IML method can have different levels of stability in different data sets, regardless of the data sizes. For example, in classification, logistic ridge regression has an AO of 0.806 in PANCAN data, but its AO score is only 0.224 in the DNase data, where both of the data are high dimensional. For a single data set, different IML methods result in different stability levels. In the DNase data, interpretations from permutation with RF are quite consistent with over AO of over 0.9, but the AO scores of other IML methods are all below 0.3. We also notice that most of the methods have less reliable interpretations in large P data sets. And in general, RF is a better choice for data with a moderate \# observation/\# feature ratio. Tree-based and linear models usually work better than MLP methods in terms of interpretation stability. Comparing model-agnostic methods, permutation with RF and logistic ridge as base models better performance on large N data. On average, the permutation method is more consistent than Shapley values. 

The heatmap in Figure~\ref{fig:f2}(D) shows the average predictive accuracy on test sets of each ML method on each data, and the accuracy scores are similar across the ML methods. None of the methods have high accuracy scores in Madelon, Amphibians, and Theorem, which are more difficult to classify. All of the methods have over 0.9 accuracies in Author, Call, and Bean data sets. Linear models and tree-based models result in high accuracy score of over 0.9 in most of the data sets, except the decision tree is less accurate in most of the high dimensional data. MLP is also relatively accurate in most of the data sets. 
 
\begin{figure}[h!]
\centering
\includegraphics[width=\linewidth]{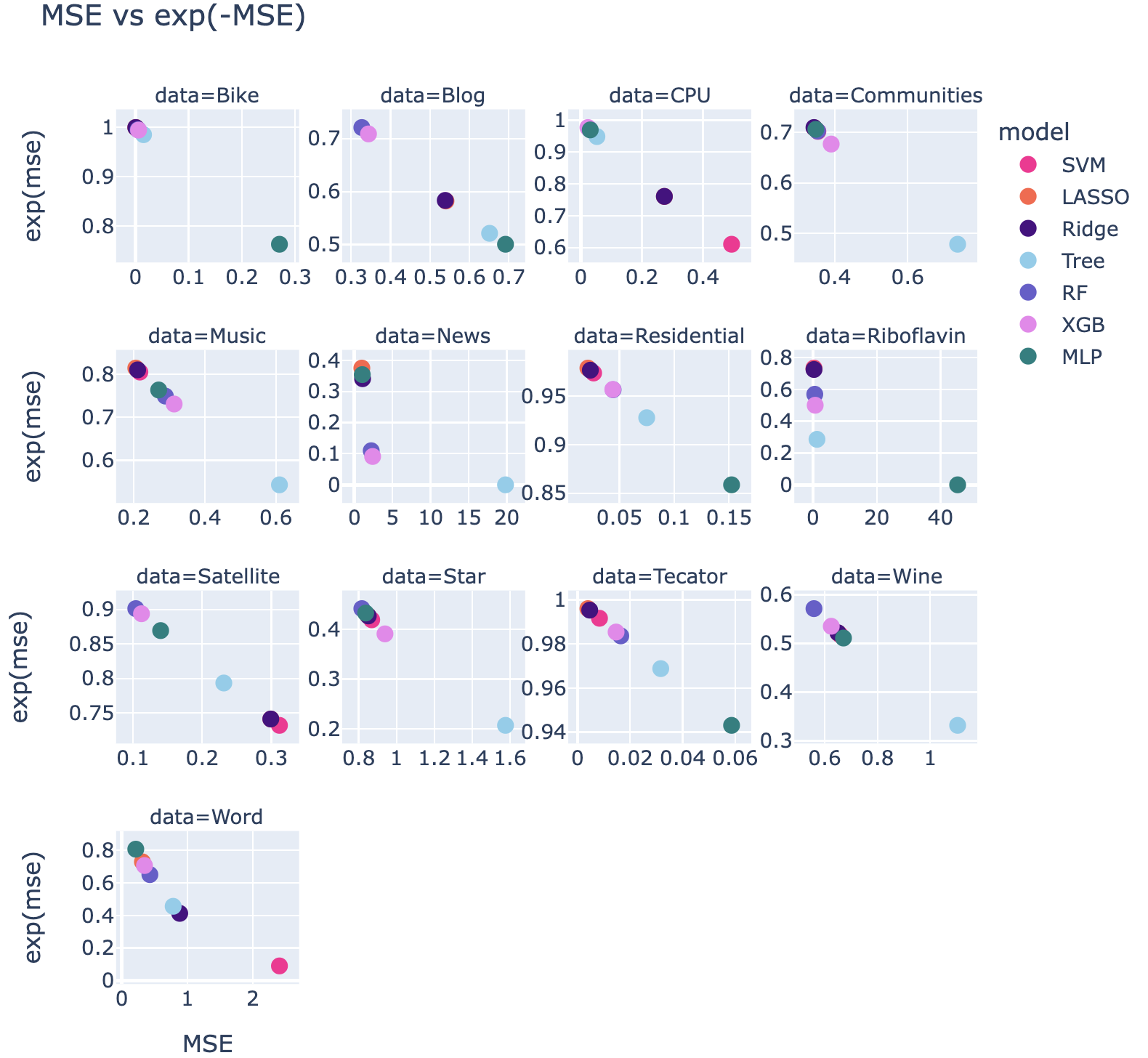}
\caption{Scatterplot of the exponential of negative MSE and MSE, colored by machine learning models of each regression data set. The transformation is almost linear for smaller MSE values and the transformed accuracy scores are within the proper range when the MSE is too large. }
\label{fig:mse}

\end{figure}

\paragraph{Regression} From Figure~\ref{fig:f3}, we see that as compared to classification results in Figure~\ref{fig:f2}(A), the stability scores of linear and tree-based models are higher for the mid-size data sets including Music, Wine, CPU, and Bike. Also, the MLP-based methods and Shapley value-related methods also have better stability than in classification. The heatmap in Figure~\ref{fig:f3}(D) shows the average predictive accuracy on test sets of each ML method on each data. To be consistent with other tasks, in regression, we transform the prediction loss measurement to an accuracy metric by taking the exponential of negative MSE between the predictions and true response in the test sets. We validate our choice of accuracy transformation in regression through Figure~\ref{fig:mse}, which shows that the transformation is almost linear for smaller MSE values and the transformed accuracy scores are within the proper range when the MSE is too large. All ML methods make accurate predictions on Tecator and Residential data sets, and less accurate predictions for the Star and News data. And they have mixed levels of prediction accuracy in Riboflavin, Word, and Blog. Combining the results in Figure~\ref{fig:f3}(A) and (D), we can still infer that models with similar accuracy may not have similar interpretations. And note that though the prediction accuracy for Wine is moderate (around 0.5), the interpretation stability is relatively high for most of the IML methods (around 0.9).

\paragraph{Clustering}
We find that a clustering algorithm can have vastly different levels of stability in different data sets. Even for a single data set, different IML methods result in different stability levels. For example, in Figure~\ref{fig:f4main}, interpretations in Religion data are quite inconsistent, except Birch has over 0.9 stability score. The heatmap in Figure~\ref{fig:f4main}(C) shows the average clustering accuracy using the true label of each clustering method on each data. It matches our expectation that a clustering algorithm can have different levels of accuracy in different data sets, due to different assumptions about the data structure. When combining results in Figure~\ref{fig:f4main}(A) and (D), we can infer that higher levels of clustering accuracy actually do lead to higher interpretation stability, as they are both based on clustering labels. However, the overall stability scores are higher than accuracy levels, as the interpretations can be consistently incorrect. For example, the clustering accuracy of HC (single) is low in most of the data sets, but it has moderate interpretation stability. With a clustering ARI of 0.472 in the Tetragonula data, HC (single) reaches an interpretation stability of 0.993. The interpretations yielded from K-Means and HC-related clustering algorithms in Iris and Tetragonla data sets are quite consistent (around 0.8), even though their clustering accuracy scores are around 0.5. The bump plot in (B) carries similar information as in the feature importance. For different data sets, the most consistent clustering algorithm is different, and there is no pattern relating to the data's \# observation/\# feature ratio. On average, the spectral (RBF) and K-Means++ generate the most consistent clustering over all data sets, and HC (single) generates the least consistent clustering labels in data splitting.


\paragraph{Dimension Reduction} 

\begin{figure}[h!]
\centering
\includegraphics[clip,width=0.7\textwidth]{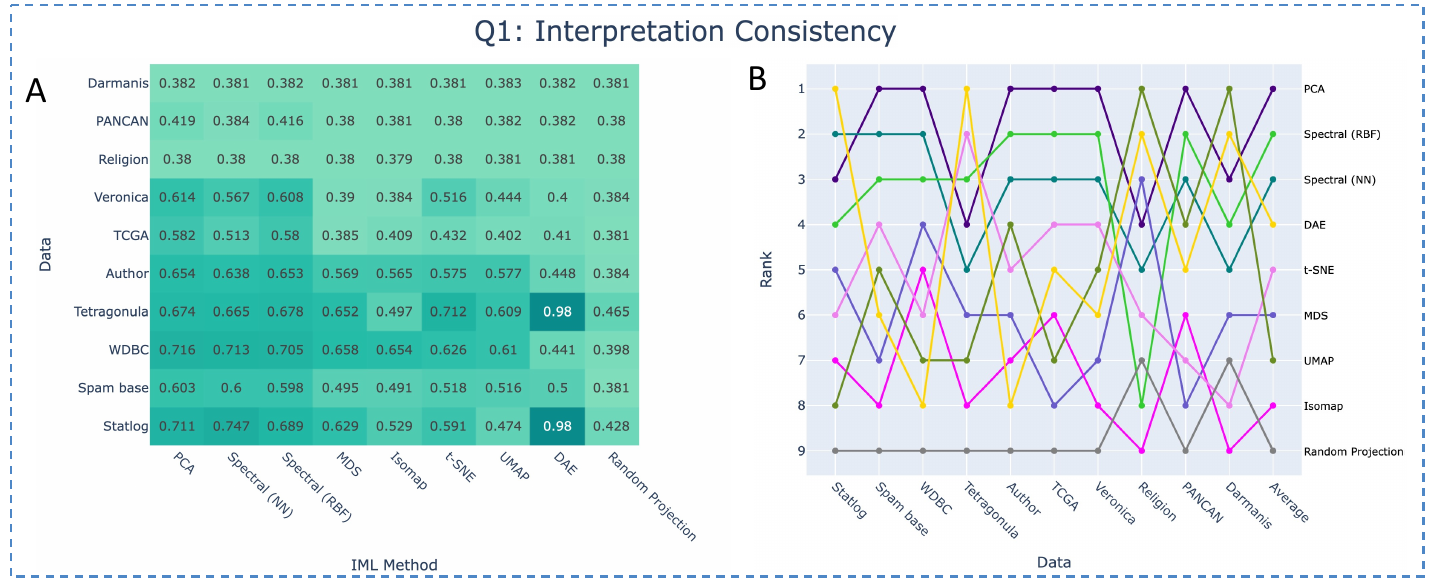}
\caption{Interpretation stability of nearest neighbors in dimension reduction IML methods with rank 2 normal noise addition with sigma = $0.15$. \textbf{A:} Heatmap of within-method interpretation stability. \textbf{B:} Bump plot of IML methods ranked by the level of interpretation stability. }
\label{fig:knn}
\end{figure}

We evaluate the interpretation stability from two perspectives in the dimension reduction tasks: clustering stability on reduced dimension and nearest neighbor stability. Figure~\ref{fig:dr}(A) and (B) show the within-method clustering label stability, using hierarchical clustering with ward linkage and Euclidean distance on the reduced dimension with rank 2. The stability is measured by the ARI metric. Plot (A) and (B) in Figure~\ref{fig:knn} show the within-method nearest neighbors stability with noise addition perturbation using normal noise with a standard deviation of 0.15, measured by the \textit{NN-Jaccard-AUC} score, as explained in Section~\ref{sec:metric} and Section~\ref{app:secmetric}. From both figures, we still have similar answers to Q1 that the interpretations are not reliable within-method. Similar to the clustering task, the accuracy and stability of clustering on the reduced dimensions demonstrate similar trends. Veronica has both relatively high clustering accuracy and stability. As Darmanis data is a single-cell RNA-seq data set, it contains a large number of zeros. Its overall performance would be better with feature selection or imputation. 

The method of random projection can be used as a baseline model, which yields the lowest accuracy and stability overall. The local methods t-SNE and UMAP work well for the bulk cell RNA-seq PANCAN data, but have poor performance in the single cell RNA-seq Darmanis data, possibly due to the sparsity of scRNA-seq data. Also, inaccurate methods/data may have consistent clustering labels. For example, TCGA data has 0 clustering accuracy for all methods, but its labeling stability is moderate. The global methods PCA and spectral clustering can result in consistent clustering labels on average, as shown in plot (B) in Figure~\ref{fig:dr}.

The nearest neighbor stability is relatively low for most methods and data sets, as shown in  plot (A) Figure~\ref{fig:knn}. Comparing the within-method stability heatmaps (A) in Figure~\ref{fig:dr} and Figure~\ref{fig:knn} and bump plots (B) in Figure~\ref{fig:dr} and Figure~\ref{fig:knn}, on the same reduced dimensions, the stability ranks of nearest neighbors are quite different from that of clustering. For example. even though the clustering labels based on DAE are not very consistent, the nearest neighbors are quite consistent with larger values, especially for the Statlog and Tetragonula data sets.  And the bump plot (B) Figure~\ref{fig:knn} shows that DAE generates the most consistent nearest neighbors, while it is the least consistent in clustering.

\subsection{Q2: Between-method stability} 

The between-method stability heatmaps in Figures~\ref{fig:f2}, \ref{fig:f3}(C) show that in classification and regression, on the same training set, the interpretations from different IML methods are not consistent with each other, while the predictions in the test sets (E) across different IML methods are quite consistent. Combining with results in (C) and (E), we can infer IML methods can generate quite different interpretations of the same data, even though their predictions on the test set are very consistent. The tree-based models and their related model-agnostic methods are relatively consistent with each other, and the same is true for linear models. 

\paragraph{Classification}

\begin{figure}[h!]
\centering
    \includegraphics[clip,width=\textwidth]{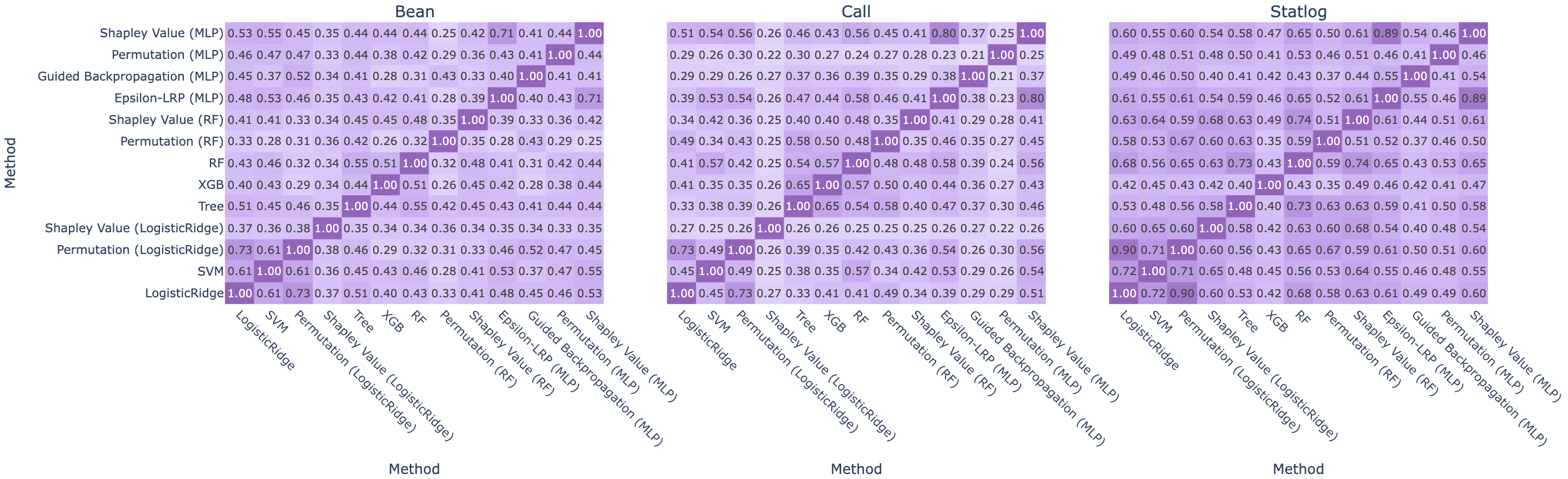}
    \caption{Detailed Results: Between-method stability Heatmap of Each Data Set.}
        \label{fig:raw_heat}
\end{figure}

 Most of the AO values in the between-method stability heatmap in (C) ranging from 0.2-0.3 in classification. The tree-based methods decision tree, RF, and XGB are relatively more consistent with each other with AO around 0.4, and permutation methods are more consistent with their base models, as the ML models are the same. Also, we notice that Epsilon-LRP is relatively consistent with Shapley value (MLP), with an AO of 0.6. The predictions in the test sets (E) across different IML methods have pairwise accuracy all greater than 0.7. The linear models SVM and logistic ridge have similar predictions with a pairwise prediction stability score of over 0.9, and the same is true for RF and XGB, which are both tree ensemble methods. In addition, though SVM and logistic ridge have pairwise prediction stability of 0.917, their interpretation stability AO score is only 0.34. Therefore, models with similar predictions may not have similar interpretations. Detailed figures of each data tell a similar story, as shown in Figure~\ref{fig:raw_heat}. 
 
\paragraph{Regression}

The overall stability scores in the regression task are higher than in classification. Epsilon-LRP is quite consistent with Shapley value (MLP) as well. The prediction stability in the middle scatterplots is measured by the exponential negative entropy, where entropy is measured by the average standard deviation of all predicted values for every single observation, which is detailed in Section~\ref{sec:metric} and Section~\ref{app:secmetric}. Most of the ML predictions are fairly consistent with each other with over 0.75 stability scores, except that the predictions of decision trees are less consistent with linear or MLP methods.

\paragraph{Clustering}
The between-method stability heatmap in Figures~\ref{fig:f4main}(C) shows that on average, the interpretations from different clustering algorithms are moderately consistent with each other with most of the ARI values around 0.5. The K-Means and K-Means++ are quite consistent with each other with an ARI of 0.812. All other methods are similarly consistent with each other, except HC(single), which is not accurate overall.

\paragraph{Dimension Reduction}

We consider the clustering label stability across different methods on the reduced dimension. The overall stability scores across different methods are moderate with most of the ARI around 0.5, as shown in the heatmap (C) in Figure~\ref{fig:dr}, similar to the clustering results. Clustering labels based on t-SNE are relatively consistent with PCA, Spectral (NN), and UMAP with ARI greater than 0.7, while PCA, Spectral (NN), and UMAP are not very consistent with each other.

\subsection{Q3: Interpretation stability vs. accuracy}

Note that under the task of clustering, we treat the clustering labels are a kind of interpretation of the underlying structure of the data set. Therefore, Q3 does not apply here as the model accuracy is also calculated by the resulting labels. 

\begin{figure}[h!]
\centering
    \includegraphics[clip,width=\textwidth]{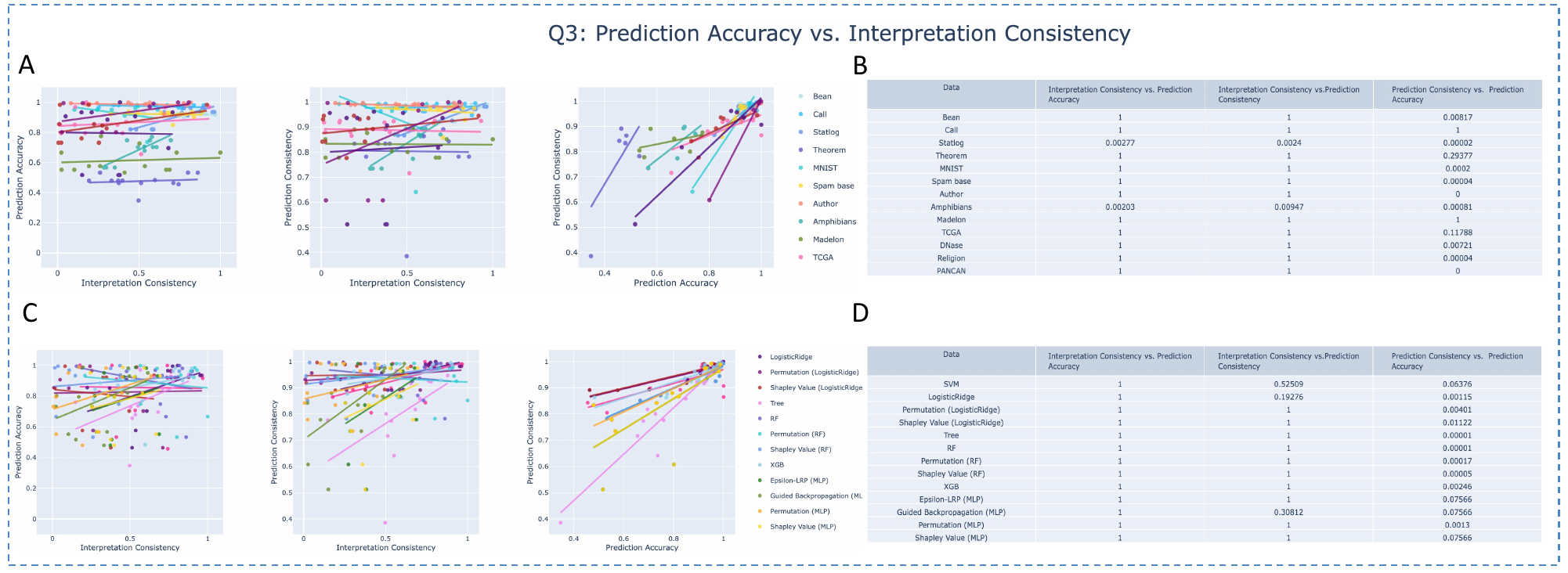}
    \caption{\textbf{A:} Summary scatterplots for each data in classification. Left: Interpretations stability against prediction accuracy; Middle: Interpretations stability against prediction stability; Right: prediction stability against prediction accuracy. \textbf{B:} P-values of fitted coefficients in classification data sets. \textbf{C: }Summary scatterplots for each IML method in classification. Left: Interpretations stability against prediction accuracy; Middle: Interpretations stability against prediction stability; Right: prediction stability against prediction accuracy. \textbf{D:} P-values of fitted coefficients in classification IML methods.    
    }
    \label{fig:class_fit}
\end{figure}

\begin{figure}[h!]
\centering
    \includegraphics[clip,width=\textwidth]{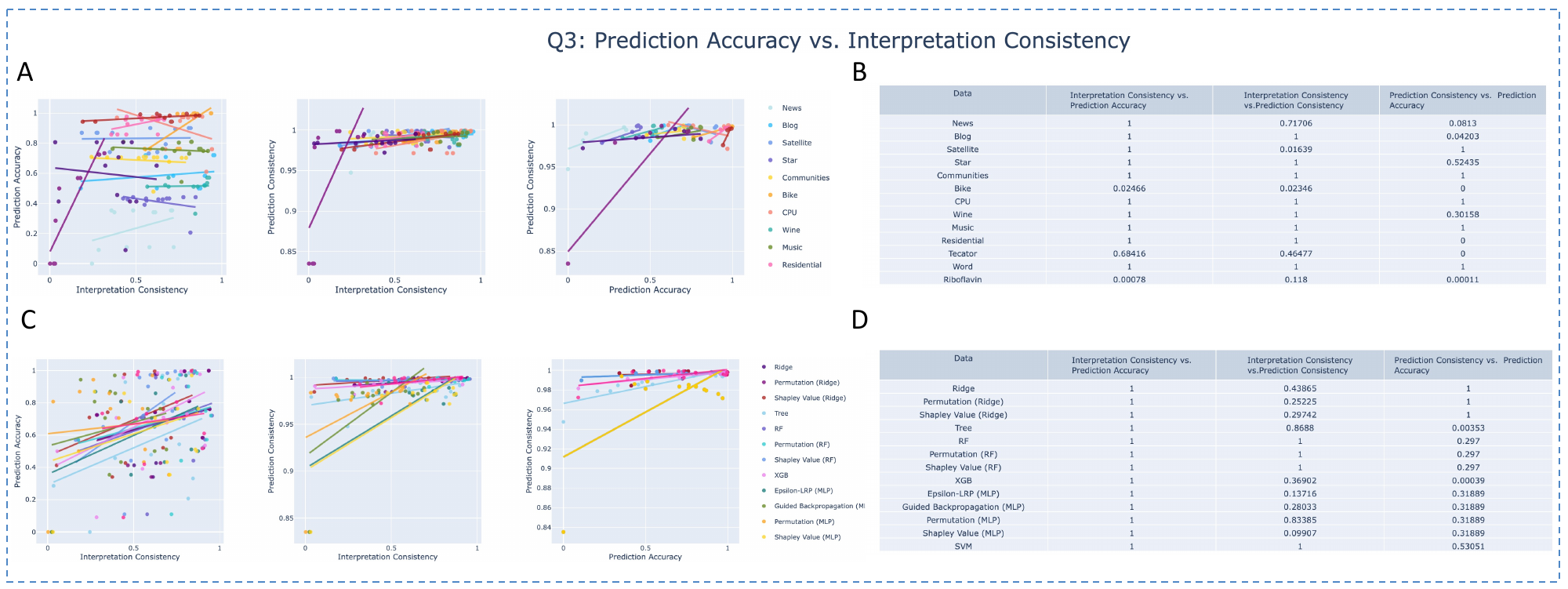}
    \caption{Answering Q3 From the Dashboard. \textbf{A:} Summary scatterplots for each dataset used in regression. Left: Interpretations stability against prediction accuracy; Middle: Interpretations stability against prediction stability; Right: prediction stability against prediction accuracy. Figure~\ref{fig:f4main} P-values of fitted coefficients in regression data sets. \textbf{C:} Summary scatterplots for each IML method in regression. Left: Interpretations stability against prediction accuracy; Middle: Interpretations stability against prediction stability; Right: prediction stability against prediction accuracy. \textbf{D:} P-values of fitted coefficients in regression IML methods.    
    }
    \label{fig:reg_fit}
\end{figure}

To explore whether predictive accuracy can be used as an indicator for interpretation stability, besides the scatterplots shown in Figure~\ref{fig:f2}, we further investigate the relationship between prediction stability and interpretation stability, as well as the relationship between prediction stability and prediction accuracy. In both Figure~\ref{fig:class_fit} and Figure~\ref{fig:reg_fit}, scatterplots (A) are aggregated over each data set, where colors represent different data sets, and scatterplots (C) are aggregated over each IML method, where colors represent different methods, all averaged over 100 repeats. In each scatter plot, we construct fitted regression lines over one data or one method. In scatterplots (A) and (C), the left scatterplots are the same as Figure~\ref{fig:f2}, which reveals the relationship between prediction accuracy and interpretation stability; the middle scatterplots aim to illustrate the relationship between prediction stability on the test set and the interpretation stability; and the right scatterplots explore the relationship between prediction stability and the prediction accuracy. We also visualize the relationships by fitting regression lines in each plot, either aggregated over data or IML methods. Lines with significant slope coefficients indicate there might be a stronger relationship between the values on the y-axis and the values on the x-axis, and flat lines indicate fewer correlations. To quantify the relationships among the interpretation stability, prediction stability, and accuracy scatterplots with respect to each method as well as each data, we also report the table of corresponding p-values of the significance of the fitted coefficients. The tables (B) and (D) in Figure~\ref{fig:class_fit} and Figure~\ref{fig:reg_fit} quantify the relationships among the interpretation stability, prediction stability, and accuracy scatterplots with respect to each method as well as each data. Similar to the classification results, on a given data set, higher predictive accuracy from an IML method does imply higher interpretation stability, as most of the fitted lines are quite flat in the left plot of (A) in Figure~\ref{fig:reg_fit}. Only the fitted lines of two data sets have significant p-values (<0.05) after the Bonferroni correction. From the perspective of IML methods, although all of the fitted lines seem to have positive coefficients, none of them are significant with p-values equal to 1, due to high variance. Therefore, for a given method, higher accuracy on one data set does not imply higher interpretation stability. None of the classification IML methods result in significant connections between interpretation stability and predictive accuracy. 

The prediction stability in the middle scatterplots is measured by the purity of the predicted label for every single observation, over the multiple train/test splits, which is detailed in Section~\ref{sec:metric}. As shown in the second columns, on a given data set, consistently predicted estimates do not imply higher interpretation stability. And for a given method, higher prediction stability on one data set does not imply higher interpretation stability as there is no significant correlation for each method as well, though all of the MLP-related methods have noticeably positive coefficients. Therefore, stability in prediction does not imply stability in feature importance interpretations. 

\paragraph{Classification}

From the left scatterplots for classification, only the fitted line of data Statlog and Amphibians have significant p-values (<0.05) after Bonferroni correction, as shown in the p-value table (C) of Figure~\ref{fig:class_fit}. From the perspective of IML methods, though there are several methods that have positive coefficients (logistic ridge, tree, XGB, and Epsilon-LRP (MLP), the coefficients are not significant with p-values equal to 1 after Bonferroni correction, because of the high variance of the results from different data sets. 

The prediction stability in the middle scatterplots is measured by the purity of the predicted label for every single observation, over the multiple train/test splits, which is detailed in Section~\ref{sec:metric}. The fitted lines in both of the middle scatterplots are similar to the left ones. As shown in the second columns in table (B) and (D) of Figure~\ref{fig:class_fit}, on a given data set, consistently predicted estimates do not imply higher interpretation stability. Only the predictive stability and interpretation stability are significantly correlated in Statlog and Amphibians data sets. And for a given method, higher prediction stability on one data set does not imply higher interpretation stability as there is no significant correlation for each method as well, though SVM, logistic ridge, tree, XGB, and Epsilon-LRP (MLP), and guided backpropagation (MLP) have noticeably positive coefficients. Therefore, stability in prediction does not imply stability in feature importance interpretations as well in classification. The right figures plot the prediction stability against predictive accuracy in test sets. We would expect the coefficients in the right scatterplot to be significant as more accurate predictions should be more consistent with each other. As shown in the third columns in table (B) and (D) of Figure~\ref{fig:class_fit}, all of the classification data sets except Call and TCGA have significant coefficients, and the same is true for IML methods except SVM and some of the MLP-based methods. The results indicate in most cases, prediction stability is significantly correlated with predictive accuracy, which matches our expectations as accurate predictions are similar. 

\paragraph{Regression}
Similar to the classification results, on a given data set, higher predictive accuracy from an IML method does imply higher interpretation stability, as most of the fitted lines are quite flat in the left plot of plot (A) in Figure~\ref{fig:reg_fit}. Only the fitted line of data Bike and Riboflavin have significant p-values (<0.05) after Bonferroni correction, as shown in the p-value table (B) of Figure~\ref{fig:reg_fit}. Note that in Riboflavin data, MLP-related methods perform poorly in terms of both prediction accuracy and interpretation stability with almost 0 values, which causes a significant p-value between predictive accuracy and interpretation stability. From the perspective of IML methods, although all of the fitted lines seem to have positive coefficients, none of them are significant with p-values equal to 1, due to high variance. Therefore, for a given method, higher accuracy on one data set does not imply higher interpretation stability. None of the classification IML methods result in significant connections between interpretation stability and predictive accuracy. 

The prediction stability in the middle scatterplots is measured by the purity of the predicted label for every single observation, over the multiple train/test splits, which is detailed in Section~\ref{sec:metric}. As shown in the second columns in table (B) and (D) in Figure~\ref{fig:reg_fit}, on a given data set, consistently predicted estimates do not imply higher interpretation stability. Satellite and Bike data sets have significant coefficients between predictive stability and interpretation stability. And for a given method, higher prediction stability on one data set does not imply higher interpretation stability as there is no significant correlation for each method as well, though all of the MLP-related methods have noticeably positive coefficients. Therefore, stability in prediction does not imply stability in feature importance interpretations in regression. Though we believe more accurate predictions should be more similar to each other, in the case of regression, the prediction stability and accuracy are not significantly correlated in some of the data sets (right plot of (A) in Figure~\ref{fig:reg_fit}) and for most of the IML methods (right plot of (C) in Figure~\ref{fig:reg_fit}). 

\section{Dashboard}
\label{app:dashboard}


%
We have provided summary figures that demonstrate only a small part of the empirical results in this supplement material. In total, our empirical study includes over 50 IML methodologies, 33 data sets, over 15 perturbation variations with different noise types and levels, 8 stability metrics, the number of top features from 1 to 30 in feature importance, and the number of ranks 2, 5 and 10 in dimension reduction. We have run over of 250,000 repeats throughout this empirical study in total, with 100 repeats for each feature importance method and 50 repeats for each unsupervised method.  
More summary figures and detailed figures of more IML methods, benchmark data sets, stability metrics, and different perturbation settings can be found in our well-developed interactive dashboard \url{https://iml-reliability.herokuapp.com/home}. 

However, researchers may ask the questions that, what if they develop an IML methodology, have a new data set to explore, or are interested in some other stability metrics other than the ones we provide? Can they evaluate and compare such reliability results under the same framework? Therefore, in addition to providing the full results for researchers and machine learning practitioners to explore, we also build a powerful function that allows users to easily upload, evaluate, and compare their own reliability results. Such results can be generated with the reliability Python package we develop, which can be found at \url{https://github.com/DataSlingers/IML_reliability}.

\subsection{App interface and interactive options}
\begin{figure}[h!]
\centering
    \includegraphics[clip,width=0.7\textwidth]{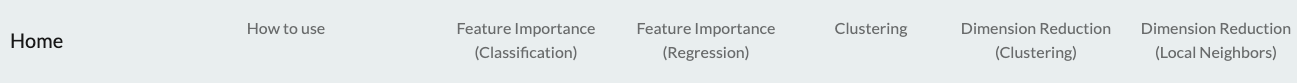}
    \caption{Dashboard navigation bar.}
    \label{fig:dash1}
\end{figure}

The dashboard consists of seven pages (Figure~\ref{fig:dash1}) including a home page ("Home"), an instruction page ("How to use"), and 5 results pages: "Feature Importance (Classification)", "Feature Importance (Regression)", "Clustering", "Dimension Reduction (Clustering)" and "Dimension Reduction (KNN)". All of the results are computed prior to dashboard visualization.

\begin{figure}[h!]
\centering
    \includegraphics[clip,width=\textwidth]{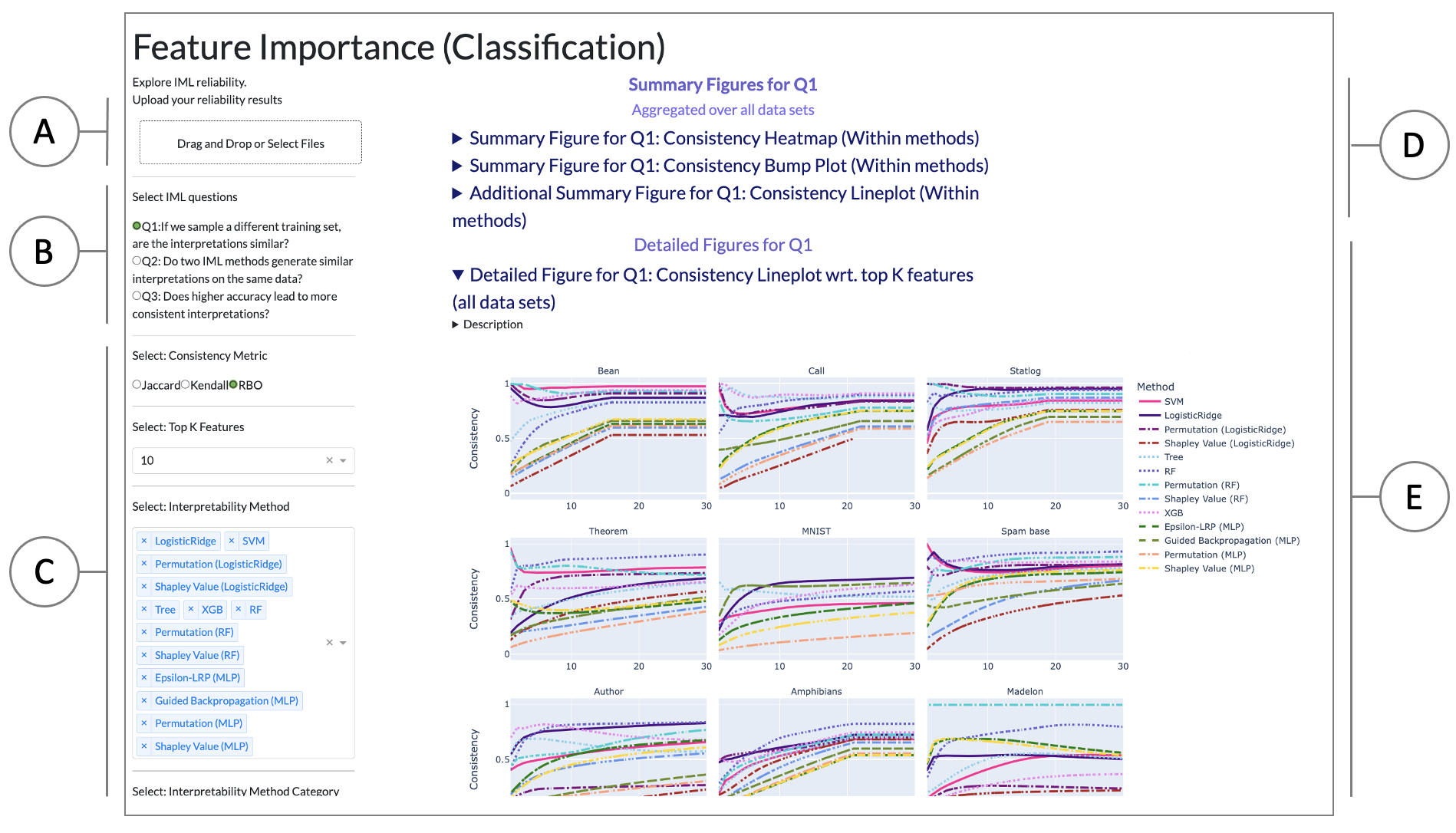}
    \caption{Screenshot of a list components of in feature importance (classification) page of the dashboard. (A): The user can upload their own reliability results for visualization using the drag-and-drop dashboard builder; (B): interactive choices to display figures, according to the specific IML questions; (C): parameters filters and selections of IML methods and data; (D): summary figures of IML results over all data sets; (E): detailed figures of IML results of each data set. }
    \label{fig:dash}
\end{figure}

The home page gives an overview of the dashboard, introducing details of the IML tasks, reliability tests, stability metrics, and data sets. The instruction page walks through the usage of the dashboard, including the layout of results pages, figure display options, data uploading function, and the selection function of data sets, IML methods, stability metrics, perturbation methods, etc. It also includes explanations of all the summary and detailed figures, and the related reliability questions. The five results pages have similar page layouts. Each page consists of two components: toolbars for interactive implementation choices, and corresponding plotly results figures, as shown in the screenshot of the feature importance (classification) page of the dashboard (Figure~\ref{fig:dash}). Our IML study focuses on three questions: 
\begin{enumerate}
    \item Q1. If we sample a different training set, are the interpretations similar?
    \item Q2. Do two IML methods generate similar interpretations on the same data?
    \item Q3. Does higher accuracy lead to more consistent interpretations?
\end{enumerate}
We construct results figures based on which question they aim to answer and the users can explore different summary and detailed figures corresponding to different questions by selecting the options in (B). 

In classification and regression tasks, we conduct reliability tests on 13 data sets with 20 IML methods prior to dashboard visualization. The default figures show the results of 13 IML methods of all data sets, including model-specific methods SVM, (Logistic) Ridge, decision tree, RF, XGB, epsilon-LRP (MLP), guided backpropagation (MLP), and model agnostic methods permutation and Shapley values, applied with base models (Logistic) Ridge, RF, and MLP, respectively. We set the default number of top K features to be evaluated as 10 and the reliability metric as AO because it puts higher weights on the top-ranked features. The details of parameters can be found in Section~\ref{app:secmetric}. Some may argue that the feature ranks can be similar if $K$ is too large and includes noise features. In the dashboard, users can explore how the reliability changes with different numbers of top features by varying $K$ (1-30) using the dropdown components. And results with reliability metrics Jaccard and Kendall's Tau are available by changing the "stability Metric" choices, which will trigger interactive actions to update the figures. 

On the clustering task results page, we present reliability results on 10 popular clustering methods with 14 benchmark data sets. We calculate the results with two kinds of perturbation methods: data splitting and noise addition. The default figures show the results with data splitting with ARI metric for both interpretation stability and clustering accuracy. Users can explore the IML reliability with noise addition by changing the "Perturbation Method" to "Noise Addition", and two more parameter selections will show up when "Noise addition" is selected: type of added noise (normal or Laplace) and the noise level measured by the variance of added noise (sigma). The dashboard will update the figures based on data with $N(0,0.5)$ noise addition. Since some IML methods may be more sensitive to noise than others, users can explore how the stability as well as accuracy change when the noise level increases. And some may be curious about whether the types of noise would make a difference to the reliability results, so we also provide results with noise generated by Laplace distribution. 

We have two pages for the reliability results of dimension reduction methods, examining the clustering stability and the stability of local neighbors on the reduced dimensions, respectively. The "Dimension Reduction (Clustering)" page is similar to the "Clustering" page in terms of options and layouts. We have one additional parameter to choose from in dimension reduction tasks: the rank of reduced dimension. As researchers often use rank 2 in data analysis, we start with the clustering stability results with data splitting perturbation using hierarchical clustering applied on the top two reduced dimensions. As the optimal ranks that can explain most of the variance can be different in data sets, users can change the "Dimension Rank" option to visualize the differences. Also, the performance might also depend on the clustering algorithm, so we also provide results using K-Means clustering to explore. Still, we provide reliability results with noise addition, either generated from the normal distribution or Laplace distribution, and the users are free to explore how the stability changes with different levels of noise added by varying "sigma" values. 

The "Dimension Reduction (KNN)" page shows the results of local neighbor stability in reduced dimensions. We develop a metric to measure the neighborhood stability by \textit{NN-Jaccard-AUC} score, as explained in Sections~\ref{sec:metric} and \ref{app:secmetric}. In this task, we only demonstrate results to answer Q1: If we sample a different training set, are the interpretations similar, as our metric mainly focuses on within-method stability and there is no ground truth to calculate accuracy. Still, users are free to change noise addition-related parameters and the number of reduced dimensions to evaluate the reliability of IML methods. 

In all three tasks, we include the results of all the data sets and all IML methods we used in this empirical study and at the same time provide drop-down options for users to select the specific data and IML methods of their interests. In addition, researchers may have their own data or IML methods of interest using our Python package that enables users to generate reliability results that can be integrated into the dashboard. After generating reliability results from the \textit{imlreliability} python package, users can upload the new results to visualize and compare to existing results via the drag-and-drop dashboard builder in (A). In addition, if users are interested in any other methods to measure interpretation stability, our package also allows them to try out their metrics.


\begin{figure}
\centering
    \includegraphics[clip,width=0.3\textwidth]{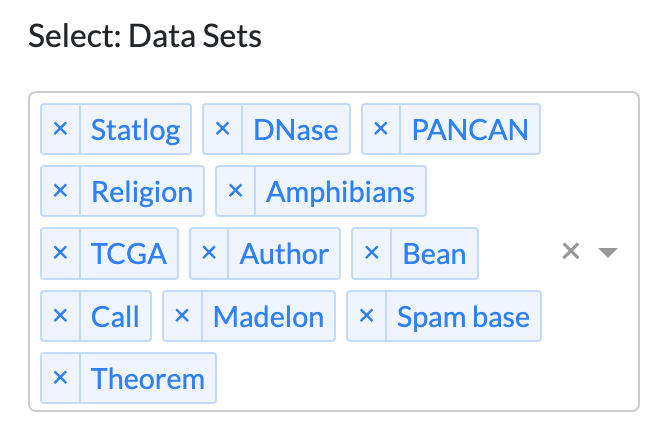}
    \caption{Example of Preloaded Datasets Within the Dashboard.}
    \label{fig:dash3}
\end{figure}

\subsection{Interactive Figures}

The interactive figures in our dashboard are organized by the three types of IML questions they aim to address, and users can visualize the results by selecting options in "Select IML questions". 

\label{app:results}
\subsubsection{Q1}
  
To address Q1 of the within-method stability, we design three types of summary figures to illustrate our empirical results, which are aggregated over data sets: (1) within-method stability heatmap, (2) Within-method stability bump plot, and (3) within-method stability line plot. And we have one detailed figure which shows the stability line plot with different levels of top K features/noise level/number of neighbors of all data sets.

\begin{itemize}
    \item \textbf{Summary Figures }
    \paragraph{Within-stability heatmap} We aim to address question 1 by measuring whether interpretations are consistent among repeats, within each method. As shown in Figure~\ref{fig:class_heat2}, the left summary heatmap demonstrates the average pairwise stability scores of interpretations of an IML method aggregated over 100 repeats, ranging in [0,1], with IML methods on the x-axis and data set on the y-axis, and the heatmap on the right side presents the each method's prediction accuracy on the test set, averaged over 100 repeats. Figure~\ref{fig:class_heat2} is an example of a within-method stability heatmap for classification feature importance methods. The left heatmap shows the within-method stability scores of each IML method on each data set, with higher values and darker colors indicating higher stability.  The blank cells are missing results because those IML methods' running times are over 12 hours for one repeat on those data sets. The right heatmap shows the prediction accuracy on test sets of each data, with higher values and darker colors indicating higher test prediction accuracy, including Logistic Ridge, SVM, Tree, RF, XGB, and MLP. From these two heatmaps, we can evaluate how each IML method performs cross different data sets, from large N to high dimensional data. Also, within one data set, we can compare the interpretation stability of different IML methods. Comparing the accuracy heatmap with the stability heatmap, we can see that even though some IML methods can generate accurate test predictions, their interpretation of important features can be inconsistent and unreliable. Users can also get more specific information by hovering over the cell of interest.

\begin{figure}[h!]
\centering
    \includegraphics[clip,width=\textwidth]{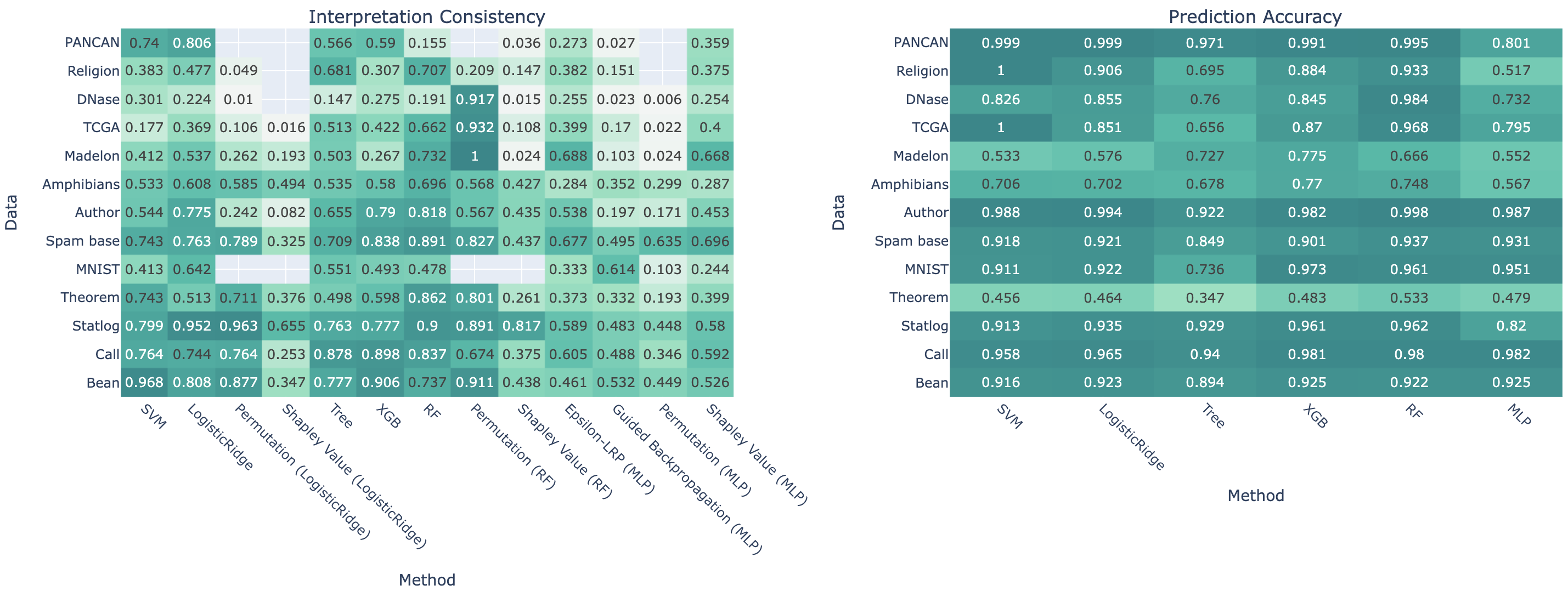}
    \caption{Summary Figure: Stability Heatmap of IML methods in Classification}
    \label{fig:class_heat2}
\end{figure}

    \paragraph{Within-stability line plot } In addition to the stability heatmap, we demonstrate the same results by plotting a line plot, with the x-axis being the data sets, ordered by \# observation/ \# feature ratio, and the y-axis being the stability score of this task, ranging in [0,1]. The left summary line plot demonstrates the average pairwise stability of interpretations of an IML method aggregated over 100 repeats. Different colors represent different methods. Figure~\ref{fig:class_line} is an example of a within-method stability line plot for classification feature importance methods. The right line plot shows the prediction accuracy on test sets of each data, including Logistic Ridge, SVM, Tree, RF, XGB, and MLP. Users can zoom in/out the figures or click the methods' names on the legend to hide or show specific methods. 
    
\begin{figure}[h!]
\centering
    \includegraphics[clip,width=\textwidth]{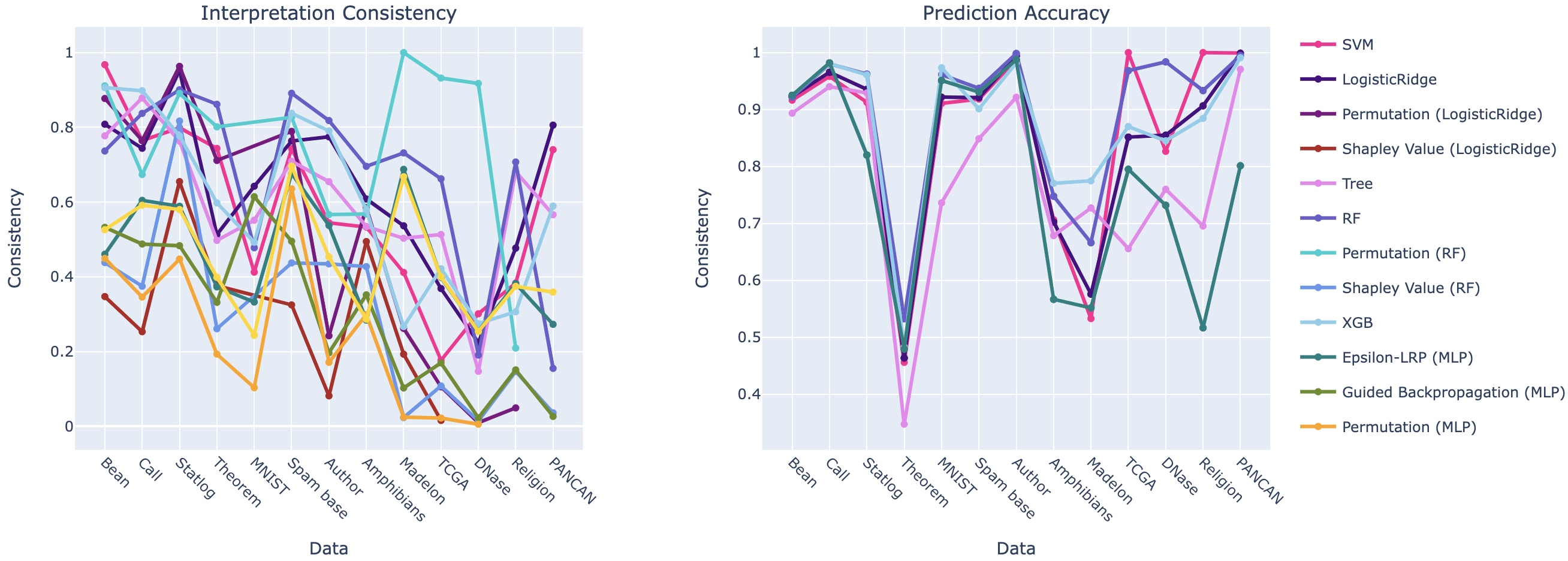}
    \caption{Summary Figure: Stability Line Plot Aggregated over Datasets)}
    \label{fig:class_line}
\end{figure}

\paragraph{Within-stability bump plot } The bump plot addresses Q1 by ranking IML methods by their stability in each data set. As shown in the example Figure~\ref{fig:class_bump}, the data sets are ordered by their \# observation/ \# ratio, with the left y-axis showing the rank of each method based on stability. We add a column of average stability over all methods at the right, and the methods of the y-axis on the right are ordered by the average stability, from the most consistent to the least consistent. There are some missing values in high dimensional data sets due to the large running time. 

\begin{figure}[h!]
\centering
    \includegraphics[clip,width=0.6\textwidth]{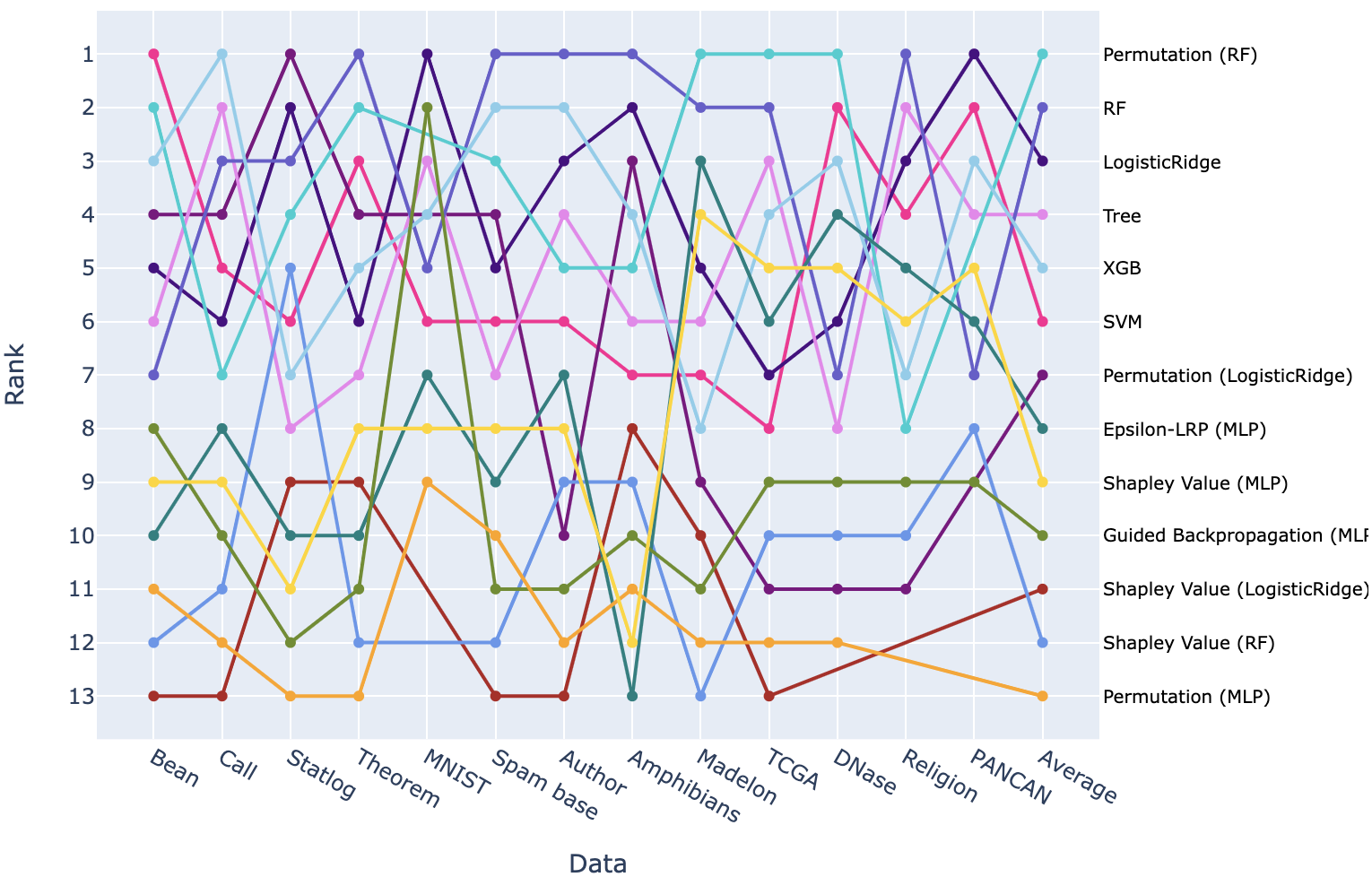}
    \caption{Summary Figure: Stability Bump Plot of Classification}
    \label{fig:class_bump}
\end{figure}
    
    \item \textbf{Detailed Figures }
    \paragraph{Detailed stability line plot } 
The detailed stability line plot shows the interpretation stability against some related parameter values of each dataset. For example, in the task of feature importance ranking, we explore the relationship between interpretation stability and the number of top features. And in the task of clustering, we aim to measure the relationship between interpretation stability and the noise level, measured by the variance of noise added to the data. In the stability of local neighbors in dimension reduction, we plot the Jaccard scores with the number of neighbors. Figure~\ref{fig:class_line_raw} illustrates the AO score against the number of top featuers $K$ in feature importance of classification methods for each data set, where colors represent different clustering methods. We are able to explore how the stability scores change with different numbers of top features and compare the trends of different IML methods in each data set. The stability scores increase with a larger number of $K$, which makes sense as more noise features are included. 

\begin{figure}[h!]
\centering
\includegraphics[clip,width=\textwidth]{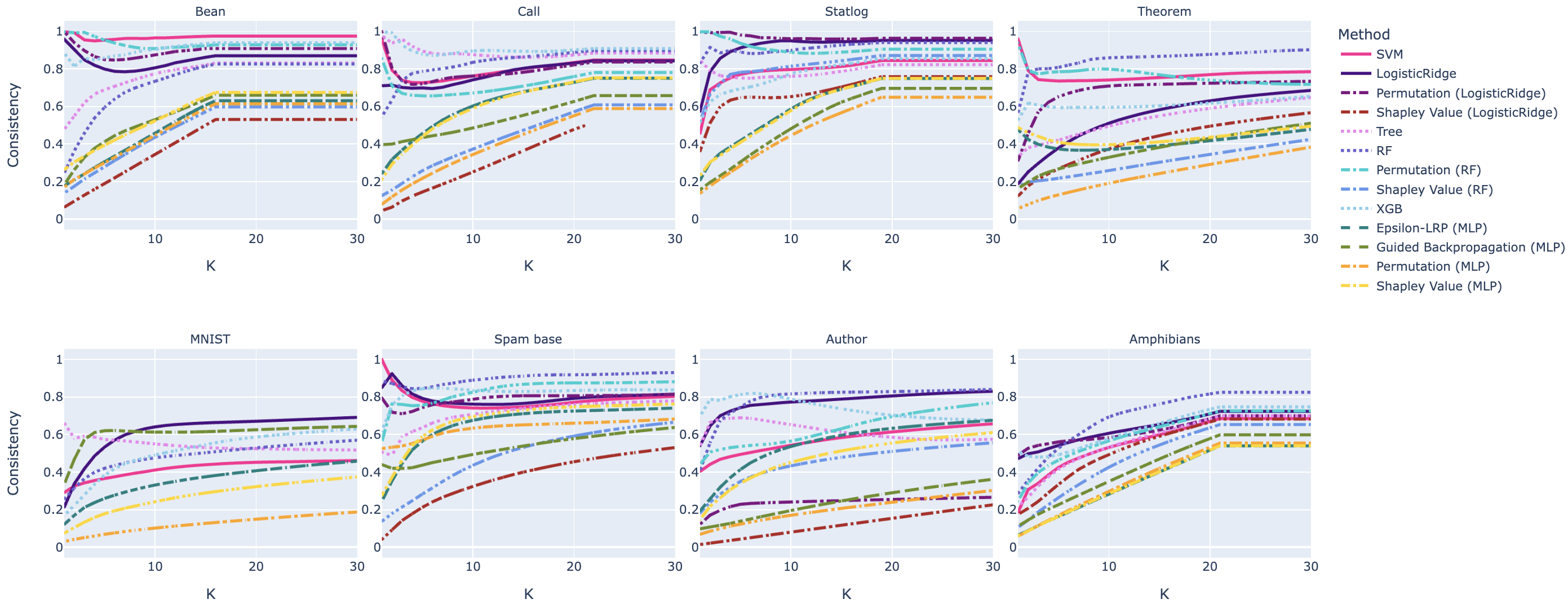}       \caption{Detailed Results: Relationship between Interpretation stability and the Number of Top Features K of Each Data Set}
\label{fig:class_line_raw}
\end{figure}
\end{itemize}

\subsubsection{Q2}

We design a summary heatmap to address Q2 of the interpretation stability between methods, which is aggregated over datasets. We also present detailed between-method stability heatmaps of all datasets separately. 

\begin{itemize}
    \item \textbf{Summary Figures }
\paragraph{Cross-stability heatmap } Among different methods, we aim to evaluate whether different methods would result in similar interpretations on the same perturbed data. As shown in Figure~\ref{fig:class_heat}, an example of a between-method stability heatmap for classification feature importance methods, the left heatmap shows the between-method average stability of interpretations obtain from each pair of IML methods, and the right heatmap shows the average stability of prediction on the test set. For instance, the cell of method LogisticRidge and method SVM represents the stability between their top K feature rankings, averaging over 100 repeats and over all data sets. The stability metric can be chosen AO, Jaccard, and Kendall's Tau. Higher values and darker colors indicate more similar interpretations between the two IML interpretations. The right heatmap shows the pairwise between-method stability of predictions on the test sets, where higher values and darker colors indicate the two methods provide more similar predictions on the same test set. 
\begin{figure}[h!]
\centering
    \includegraphics[clip,width=\textwidth]{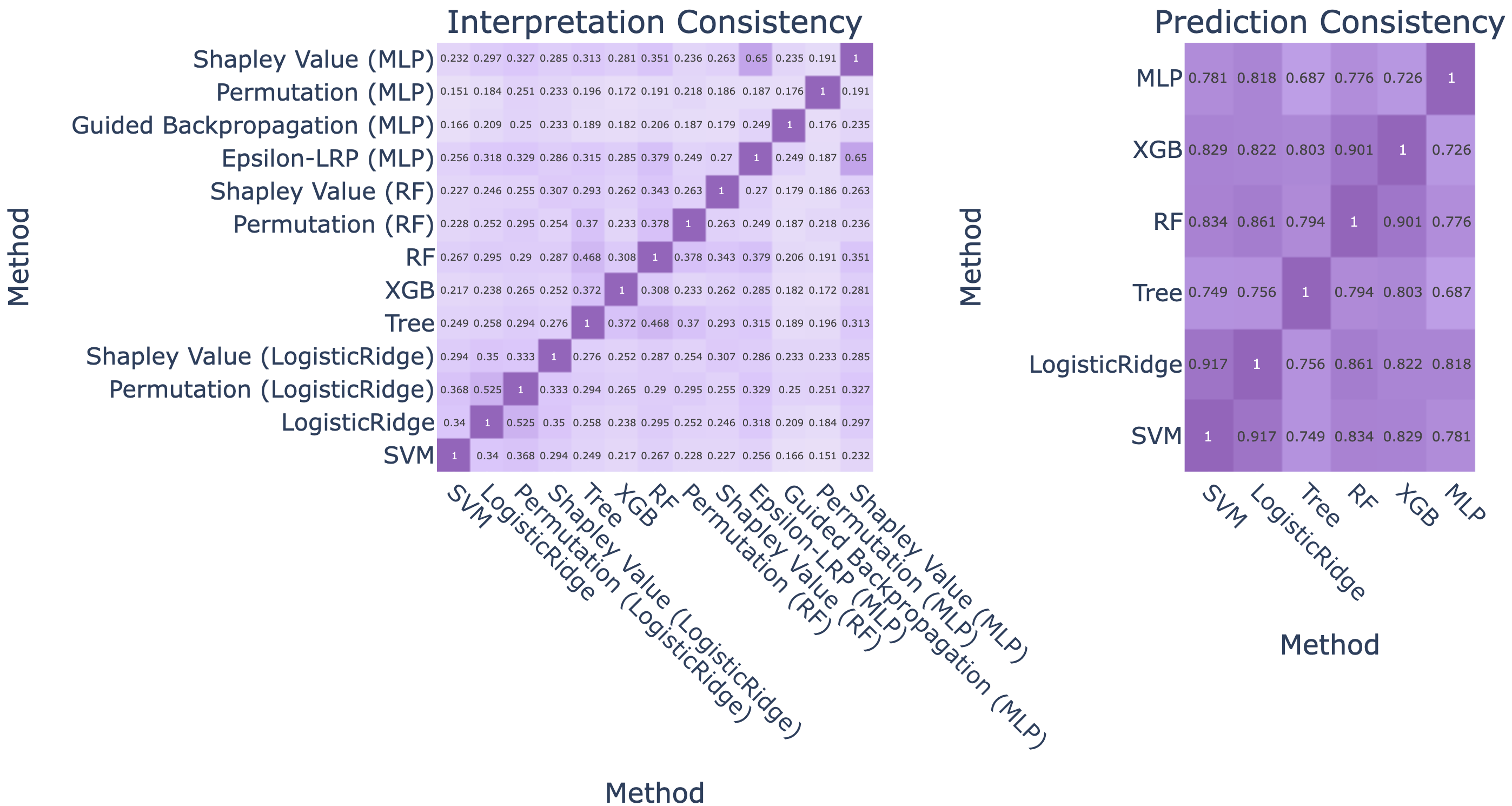}
    \caption{Summary Figure: Between-method stability Heatmap of Classification}
    \label{fig:class_heat}
\end{figure}
  \item \textbf{Detailed Figures }

\paragraph{Detailed between-method stability heatmap} The detailed between-method stability heatmap is similar to the summary heatmap figure, but it shows the between-method average stability of interpretations in each data set. Figure~\ref{fig:raw_heat} is a subset of detailed between-method heatmap in classification. Since the IML performance can vary with different data sizes or the different underlying data structures, we can explore how similar the interpretations generated by different IML methods are for each data set.

\end{itemize}

\subsubsection{Q3}

We address Q3 "Does higher accuracy leads to more consistent interpretations?" by plotting two sets of scatterplots to explore the relationships among interpretation stability, prediction accuracy, and prediction stability. These figures demonstrate the detailed scatter plots of interpretation stability and prediction accuracy of each data set.

\begin{itemize}
    \item \textbf{Summary Figures }
\paragraph{Interpretation stability, prediction stability \& accuracy scatterplots} 

We address Q3 with six scatterplots, which explore the relationship among model prediction accuracy, prediction stability, and interpretation stability, with each dot representing one IML method on one data set. As shown in the scatterplots Figure~\ref{fig:class_fit}(A),(C), which are aggregated over each data set, or each IML method, respectively. The figures demonstrate the relationship between prediction accuracy on the test set and the interpretation stability (left), the relationship between prediction stability on the test set and the interpretation stability (middle), and the relationship between prediction stability and the prediction accuracy (right). In each scatter plot, we construct fitted regression lines over one data. The significance of the fitted coefficient implies whether there exists a strong relationship between values on the y-axis and the values on the x-axis. The relationships are also quantified by the p-values for testing the fitted coefficients significance, which is reported in the tables Figure~\ref{fig:class_fit}(B),(D). Bonferroni-corrected p-values less than 0.05 indicate significant correlations. Users can check more information by hovering over the scatterplots on the dashboard.

  \item \textbf{Detailed Figures }

\paragraph{Detailed stability \& accuracy scatterplots}  We aim to explore the relationship between model prediction accuracy and interpretation stability of each data set. The detailed Figure~\ref{fig:class_scatter_raw}) is similar to the left figure of scatterplot (A) in Figure~\ref{fig:class_fit}, but separates the scatters by data sets. With scatters representing different IML methods, we can explore how the interpretation stability related to model accuracy in each data. For example, Figure~\ref{fig:class_scatter_raw}) shows the results in classification feature importance methods, and we can find that methods with higher accuracy do not necessarily have higher stability, such as the high dimensional PANCAN, DNase, and TCGA data.  

\begin{figure}[h!]
\centering
    \includegraphics[clip,width=\textwidth]{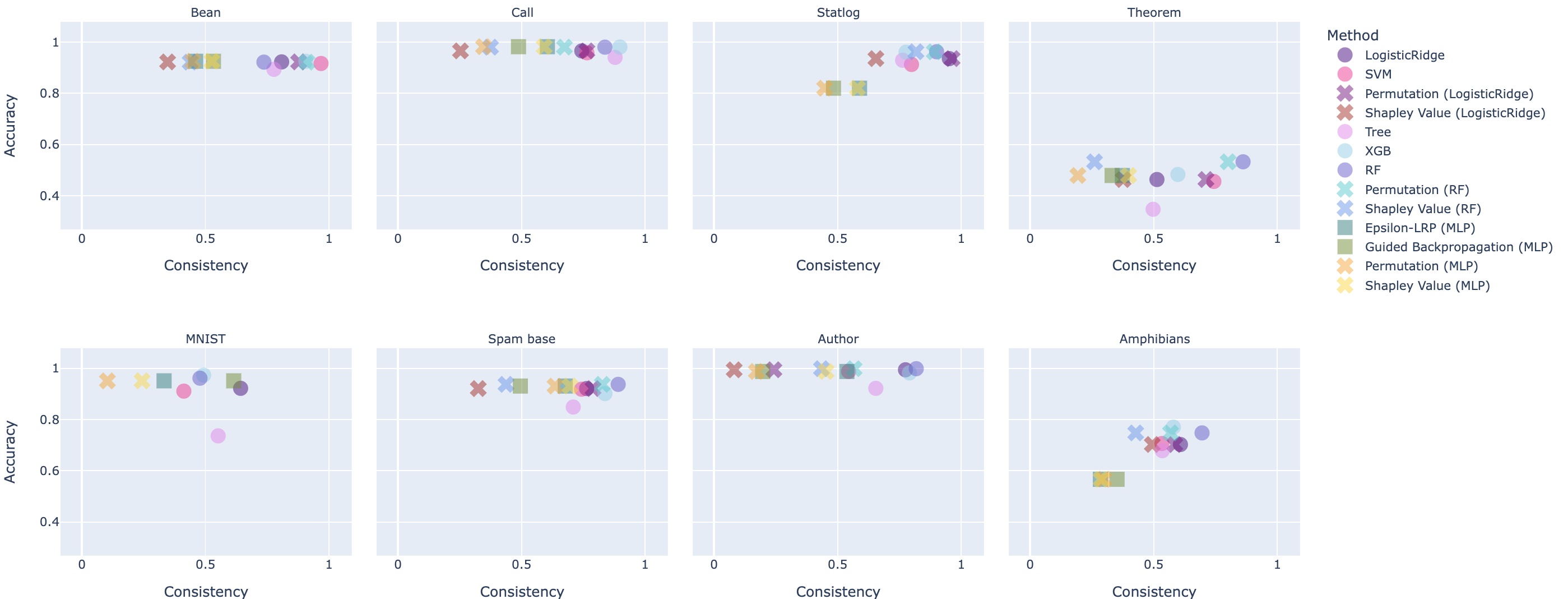}
    \caption{Detailed Results: Relationship between Interpretation stability and Predictive Accuracy of each data set}
    \label{fig:class_scatter_raw}
\end{figure}

  \end{itemize}

\bibliographystyle{abbrv}
\bibliography{ref} 
\end{document}